\documentclass[onecolumn,journal,10pt]{IEEEtran}
\usepackage[utf8]{inputenc} 
\usepackage[T1]{fontenc}    
\usepackage{bbm}            
\usepackage{microtype}      
\usepackage{amsmath,amssymb,amsthm}
\usepackage{mathrsfs}  
\usepackage{nicefrac}       
\usepackage{hyperref}       
  \hypersetup{%
	pdfmenubar=true,	    
	pdffitwindow=false,     
	pdfstartview={FitH},    
	colorlinks=true,        
	linkcolor=blue,         
	citecolor=red,          
	filecolor=blue,         
	urlcolor=blue           
  }
\usepackage{booktabs}       
\usepackage{multirow}
\usepackage{paralist}
\usepackage{enumitem}
\usepackage[subrefformat=parens,labelformat=parens]{subfig}
\usepackage{wrapfig}
\usepackage{tikz}
  \usetikzlibrary{arrows}
  \usetikzlibrary{automata}
  \usetikzlibrary{spy}
\usepackage{pgfplots}
  \pgfplotsset{width=7cm,compat=1.8}
\usepackage{algorithm,algorithmic}
\usepackage{acro}
\usepackage{xcolor}         

\usepackage[tikz]{bclogo}
\usepackage[framemethod=tikz]{mdframed}
\usepackage[many]{tcolorbox}

\definecolor{ttblue}{RGB}{91,194,224}

\usepackage{setspace} 
\usepackage[a4paper,left=1.25in, right=1.25in, top=1in, bottom=1in]{geometry}

\newcommand{\Real}{\mathbb{R}}
\newcommand{\Natural}{\mathbb{N}}
\newcommand{\X}{\mathbb{X}}
\newcommand{\Y}{\mathbb{Y}}

\newcommand{\stochastic}[1]{\mathbbm{#1}}
\DeclareMathOperator*{\argmin}{arg\,min}
\DeclareMathOperator*{\argmax}{arg\,max}

\newcommand{\Op}[1]{\operatorname{\mathcal{#1}}}
\newcommand{\ForwardOp}{\Op{A}}
\newcommand{\LossFunc}{\Op{L}}
\newcommand{\LossRec}{\LossFunc_{\X}}
\newcommand{\LossData}{\LossFunc_{\Y}}
\newcommand{\RegFunc}{\Op{S}}
\newcommand{\RecOp}{\Op{R}}
\newcommand{\x}{x}
\newcommand{\stx}{\stochastic{x}}
\newcommand{\y}{y}
\newcommand{\sty}{\stochastic{y}}
\newcommand{\e}{e}
\newcommand{\stu}{\stochastic{u}}
\newcommand{\ste}{\stochastic{e}}
\newcommand{\z}{z}
\newcommand{\stz}{\stochastic{z}}

\DeclareAcronym{ACR}{
  short = ACR,
  long = adversarial convex regularizer} 
  \DeclareAcronym{AMP}{
  short = AMP,
  long = approximate message passing} 
\DeclareAcronym{ADMM}{
  short = ADMM,
  long = alternating-directions method of multipliers}   
\DeclareAcronym{AE}{
  short = AE,
  long = auto-encoder}
\DeclareAcronym{AR}{
  short = AR,
  long = adversarial regularizer}   
\DeclareAcronym{ART}{
  short = ART,
  long = algebraic reconstruction technique}
  \DeclareAcronym{CE}{
  short = CE,
  long = consensus equilibrium} 
\DeclareAcronym{CGLS}{
  short = CGLS,
  long = conjugate gradient least squares}  
\DeclareAcronym{CNN}{
  short = CNN,
  long = convolutional neural network}  
\DeclareAcronym{CT}{
  short = CT,
  long = computed tomography}  
\DeclareAcronym{DNN}{
  short = DNN,
  long = deep neural network}  
\DeclareAcronym{FBP}{
  short = FBP,
  long = filtered back-projection}
  \DeclareAcronym{FBS}{
  short = FBS,
  long = forward-backward splitting}
\DeclareAcronym{GAN}{
  short = GAN,
  long = generative adversarial network} 
  \DeclareAcronym{GS}{
  short = GS,
  long = gradient-step} 
  \DeclareAcronym{HQS}{
  short = HQS,
  long = half-quadratic splitting} 
\DeclareAcronym{ICNN}{
  short = ICNN,
  long = input-convex neural network} 
\DeclareAcronym{LPD}{
  short = LPD,
  long = learned primal-dual} 
\DeclareAcronym{MAP}{
  short = MAP,
  long = maximum a-posteriori probability}
\DeclareAcronym{MC}{
  short = MC,
  long = Monte Carlo}
\DeclareAcronym{MCMC}{
  short = MCMC,
  long = Markov chain Monte Carlo}
\DeclareAcronym{MMSE}{
  short = MMSE,
  long = minimum mean-squared error}
  \DeclareAcronym{MMO}{
  short = MMO,
  long = maximally monotone operators}
\DeclareAcronym{MLEM}{
  short = ML-EM,
  long = expectation–maximization}   
\DeclareAcronym{NN}{
  short = NN,
  long = neural network}
  \DeclareAcronym{NETT}{
  short = NETT,
  long = network Tikhonov}
  \DeclareAcronym{PGD}{
  short = PGD,
  long = proximal gradient-descent} 
\DeclareAcronym{PnP}{
  short = PnP,
  long = plug-and-play} 
\DeclareAcronym{PSNR}{
  short = PSNR,
  long = peak signal to noise ratio}
  \DeclareAcronym{RED}{
  short = RED,
  long = Regularization-by-denoising}
  \DeclareAcronym{ReLU}{
  short = ReLU,
  long = rectified linear units}
\DeclareAcronym{SIRT}{
  short = SIRT,
  long = simultaneous iterations reconstruction technique}
\DeclareAcronym{SSIM}{
   short = SSIM,
   long = structural similarity index} 
   \DeclareAcronym{TV}{
   short = TV,
   long = total variation} 
\DeclareAcronym{VAE}{
  short = VAE,
  long = variational auto-encoder}
\DeclareAcronym{ULA}{
  short = ULA,
  long = unadjusted Langevin algorithm}
\DeclareAcronym{SGD}{
  short = SGD,
  long = stochastic gradient-descent}  

\mdfdefinestyle{mystyle}{%
  rightline=true,
  innerleftmargin=10,
  innerrightmargin=10,
  outerlinewidth=3pt,
  topline=false,
  rightline=true,
  bottomline=false,
  skipabove=\topsep,
  skipbelow=\topsep
}

\begin{document}

\title{Learned reconstruction methods with convergence guarantees}

\author{%
Subhadip Mukherjee$^{*1}$, 
Andreas Hauptmann$^{*2,3}$, 
Ozan \"Oktem$^4$, 
Marcelo Pereyra$^5$, and 
Carola-Bibiane Sch\"onlieb$^1$
\\
\small{%
  $^1$University of Cambridge, UK;
  $^2$University of Oulu, Finland; 
  $^3$University College London, UK; 
  \\ 
  $^3$KTH--Royal Institute of Technology, Sweden; 
  $^4$Maxwell Institute for Mathematical Sciences \& Heriot-Watt University, UK; 
  $^*$Equal contribution
  \\
  E-mails:%
  \href{mailto:andreas.hauptmann@oulu.fi}{andreas.hauptmann@oulu.fi}, 
  \href{mailto:sm2467@cam.ac.uk}{sm2467@cam.ac.uk}, 
  \href{mailto:ozan@kth.se}{ozan@kth.se}, 
  \href{mailto:m.pereyra@hw.ac.uk}{m.pereyra@hw.ac.uk}, 
  \href{mailto:cbs31@cam.ac.uk}{cbs31@cam.ac.uk}
 }
}

\markboth{IEEE SP Magazine Special Issue on Physics-Driven Machine Learning for Computational Imaging}%
{Shell \MakeLowercase{\textit{et al.}}: A Sample Article Using IEEEtran.cls for IEEE Journals}


\maketitle
\begin{abstract}
In recent years, deep learning has achieved remarkable empirical success for image reconstruction.
This has catalyzed an ongoing quest for precise characterization of correctness and reliability of data-driven methods in critical use-cases, for instance in medical imaging. Notwithstanding the excellent performance and efficacy of deep learning-based methods, concerns have been raised regarding their stability, or lack thereof, with serious practical implications.
Significant advances have been made in recent years to unravel the inner workings of data-driven image recovery methods, challenging their widely perceived black-box nature. 
In this article, we will specify relevant notions of convergence for data-driven image reconstruction, which will form the basis of a survey of learned methods with mathematically rigorous reconstruction guarantees. An example that is highlighted is the role of \acp{ICNN}, offering the possibility to combine the power of deep learning with classical convex regularization theory for devising methods that are provably convergent. 

This survey article is aimed at both methodological researchers seeking to advance the frontiers of our understanding of data-driven image reconstruction methods as well as practitioners, by providing an accessible description of useful convergence concepts and by placing some of the existing empirical practices on a solid mathematical foundation. 
\end{abstract}

\bstctlcite{IEEEexample:BSTcontrol}

\section{Introduction}
Image reconstruction problems are virtually ubiquitous in scientific and engineering applications; ranging from astronomy to clinical diagnosis, from tomographic imaging in 3D electron microscopy to X-ray crystallography. In such problems, an image of interest needs to be recovered from its incomplete and noisy observation. The data generation process is governed by an underlying physical process, precise knowledge of which is a prerequisite to formulate an accurate image reconstruction problem. 

It is of paramount importance, especially for critical applications such as medical imaging, to obtain image reconstructions with reliable content as they contain critical structural information about the object of interest. Moreover, in many applications of computational imaging, image reconstruction is used as a tool for scientific discovery without any ground-truth being available. Therefore, one needs to be able to rely on the correctness of the reconstruction result. Correctness and reliability of reconstruction algorithms have been traditionally provided in terms of convergence guarantees, and more specifically, in the framework of model-based variational regularization \cite{scherzer2009variational,benning_burger_2018}. Mathematically, image reconstruction is an example of an inverse problem, where research is primarily concerned with the development and analysis of mathematical theory and algorithms in a fairly general setting. Such guarantees also offer a principled framework for practitioners to control the reconstruction process. 

The emergence of deep learning and the availability of high-quality training data and computing capabilities have considerably transformed the research landscape of image reconstruction \cite{data_driven_inv_prob,lucas2018using}. 
Image quality has improved to a considerable extent as compared to the classical model-based techniques, both qualitatively and quantitatively, driven by task-specific image data sets and sophisticated machine learning algorithms. Some concerns have been voiced nevertheless, pointing out the lack of adequate comprehension of what happens within the learned reconstruction methods, putting forward the argument that a more thorough understanding is needed for reliable utilization of these techniques.
Consequently, in parallel with the ongoing enterprise of designing more efficient and better-performing data-driven reconstruction approaches, researchers have begun to investigate theoretical properties of these methods for image reconstruction. 

A specific question is whether one can provide reconstruction guarantees and convergence results for deep learning-based methods. 
This question is important to theoreticians and practitioners alike, since successful answers would place empirically well-performing methods, often based on heuristics, on a rigorous theoretical foundation. 
Nevertheless, the notion of \emph{theoretical guarantees} is imprecise and can have different meanings depending on the specific context.

This survey attempts to define different notions of convergence within the realm of image reconstruction, elaborate on their practical implications, and present an overview of recent deep learning-based image reconstruction methods that fit into the different notions of convergence. 
We argue that many of the recently proposed deep learning-based approaches come with more theoretical backing than it is often communicated and are, in fact, not as much a black-box as they are generally referred to be.
In particular, identifying the connections between model- and data-driven methods will help unite the two seemingly disparate paradigms and broaden our knowledge of this exciting new line of research. 

The article is organized as follows. In Sec.~\ref{sec:math_foundation}, we provide the mathematical preliminaries for inverse problems and explain different training strategies for data-driven image reconstruction. This section facilitates precise characterization of different convergence notions, which appears in Sec.~\ref{sec:convergence_notions}, in a rigorous yet accessible manner. The ideas of convergence considered in this article are primarily derived from the classical regularization, convex analysis, and statistics literature. Sections ~\ref{sec:provable} and \ref{sec:provable_bayesian_methods} provide a review of recent notable data-driven methods that come with convergence guarantees introduced in the preceding section, in the classical functional-analytic and Bayesian settings, respectively. Our survey is not exhaustive by any means, in that it excludes approaches that are based on heuristics and are not provably convergent (except for a few pioneering methods that inspired new lines of research). It is also worth emphasizing that the methods chosen for review are not necessarily the best-performing methods empirically, but they are more interpretable in the classical sense and hence more transparent as compared to competing techniques with possibly superior numerical performance. Finally, we present a summary and make some concluding remarks in Sec.~\ref{sec:conclusion}.

\section{Mathematical foundations}
\label{sec:math_foundation}
In order to characterize what convergence and reconstruction guarantees mean for an image reconstruction problem, we need to mathematically formulate the reconstruction task. 

In the traditional deterministic \emph{functional analytical} setting, this is viewed as solving an operator equation. More precisely, one seeks to find $\x^* \in \X$, which is unknown but deterministic, from measured data $\y \in \Y$ under the measurement model $\y = \ForwardOp\x^* + \e$, where $\e \in \Y$ denotes observation error.
Here, $\ForwardOp \colon \X \to \Y$ (\emph{forward operator}) models how an image gives rise to data in the absence of observation error and is typically derived from a careful modeling of the involved physics. In what follows, we focus on the linear setting where $\ForwardOp$ is a linear operator that can be represented by a matrix for discretized images. As we discuss convergence results for the linear setting, we will comment on extensions to the nonlinear case wherever appropriate.\\ 
The spaces $\X$ and $\Y$ can be fairly general and potentially infinite-dimensional function spaces. However for the purpose of our exposition, it suffices to consider them as finite-dimensional vector spaces endowed with an inner-product and norm (in particular, subsets of Euclidean spaces). Given the forward model, an estimate of the desired image $\x^*$ is obtained by formulating a \emph{reconstruction method}, represented by a mapping $\Op{R} \colon \Y \to \X$. In the functional analytic setting, this corresponds to a regularized inverse of $\ForwardOp$.

\emph{Bayesian inversion} extends the above by modeling the true (unknown) image $\x^* \in \X$ as a realization of an $\X$-valued random variable $\stx$ \cite{kaipio2006statistical}. The distribution of $\stx$, known in the literature as the {prior} distribution, constitutes a statistical model for images in $\X$ that encodes desired and expected properties about the solution. For example, in the context of limited-angle CT chest imaging, $\stx$ would represent our expectations about what a CT chest image looks like at a population level in noise-free high-resolution conditions, with $\x^*$ understood as a realization of $\stx$ stemming from performing a CT scan for an individual in that population. The measured data $\y \in \Y$, acquired by the scanner in limited-angle conditions and corrupted by noise, is regarded as a realization of a $\Y$-valued random variable $\sty$ conditioned on $\stx = \x^*$, with the two random variables being related by the forward model $\sty =\ForwardOp\stx+ \ste$, where $\ste$ is a random variable for the measurement noise. Given $\y$, Bayesian image reconstruction becomes a statistical inference task underpinned by the conditional distribution of $(\stx | \sty = \y)$, commonly known as the \emph{posterior} distribution of $\stx$. Bayesian estimators stem from seeking to optimally summarize this posterior distribution as a single point in $\X$, with different optimality criteria leading to different Bayesian estimators \cite{Pereyra2019}.

A key challenge in both the functional analytical and the Bayesian frameworks is to handle the \emph{ill-posedness} of inversion, which arises from the fact that the forward operator $\ForwardOp$ in practical inverse problems is either under-determined or poorly-conditioned with an unstable inverse. This leads to non-uniqueness and instability of reconstruction, meaning that many candidate images explain the measured data even in the absence of noise, or a small amount of noise in the data results in large changes in the recovered image. One can circumvent ill-posedness by introducing a \emph{regularization} to stabilize the reconstruction, for instance by enforcing certain regularity conditions, such as smoothness. In Bayesian inversion, regularization is often achieved by selecting a prior distribution on images, which assigns low likelihood to images with unwanted features. In both settings, regularization involves a handcrafted a model that encodes prior knowledge as well as expected properties of the solution.




\subsection{Model-based reconstruction}\label{sec:modelbasedRec}
The design of regularized reconstruction methods is traditionally based on the functional analytic view.
Early approaches sought to provide an analytic pseudo-inverse to the forward operator (\emph{direct regularization}). An example is the \ac{FBP} method for tomographic image reconstruction, which regularizes by recovering the band-limited part of the image. A drawback of such an approach is that it is problem-specific, and does not generalize to a different forward operator.

Thus, it is desirable to formulate a general class of reconstruction methods that allow us to replace the forward operator in a plug-and-play manner. 
This leads to \emph{variational models}, where the reconstruction task is formulated as a minimization problem of some penalty function $\Op{J} \colon \X\times\Y\to\Real$ that ensures data-consistency and incorporates a regularizer. More precisely, such a penalty function typically takes the form $\Op{J}_{\theta}(\x,\y) := \LossData\bigl( \ForwardOp\x,\y \bigr) +  \RegFunc_{\theta}(\x)$, where $\LossData \colon \Y\times\Y\to \Real$ quantifies consistency in data space $\Y$ and the regularizer $\RegFunc_{\theta} \colon \X \to \Real$ penalizes undesirable solutions. The data consistency term is often taken to be the least-squares loss $\LossData\bigl( \ForwardOp\x,\y \bigr) := \|\ForwardOp\x-\y\|_2^2$, which, if minimized without regularization, leads to overfitting the measurement noise. This underlines the need of including a regularization term $\RegFunc_{\theta}$ with possible hyper-parameters $\theta\in\Real^d$, which allows to encode prior knowledge about desirable solutions, such as sparsity assumptions \cite{benning_burger_2018, lasso_rt_retrospective, donoho_cs_transactions_IT, candes_cs_spm}. 
The reconstruction method $\RecOp_{\theta} \colon \Y \to \X$ with $\theta \in \Real^d$ is now defined as the solution operator for the minimization problem
\begin{equation}\label{eq:VarReg}
  \RecOp_{\theta}(\y) \in \argmin_{\x \in \X} 
    \Op{J}_{\theta}(\x,\y)
  \quad\text{where}\quad
  \Op{J}_{\theta}(\x,\y):=\LossData\bigl( \ForwardOp\x,\y \bigr) +  \RegFunc_{\theta}(\x).
\end{equation}
The hyper-parameter $\theta$ needs to be chosen beforehand depending on the noise level in the data. A simple yet widely popular special case is to construct $\RegFunc_{\theta}(\x)$ as $\RegFunc_{\theta}(\x)=\lambda\,\mathscr{S}_{\vartheta}(x)$, where $\theta=(\lambda,\vartheta)$ with $\lambda>0$. Typically, only $\lambda$ is adjusted depending on the noise level $\delta$, while the parameter $\vartheta$ of the functional $\mathscr{S}_{\vartheta}$ is kept fixed (either hand-crafted or pre-trained on some training data set).
Solutions to \eqref{eq:VarReg} are then often computed through an \emph{iterative scheme} by (proximal) gradient-based methods, which are the basis for the unrolling techniques discussed in Sec.~\ref{sec:provable}.

Reconstructions arising from minimizing a variational potential such as \eqref{eq:VarReg} can alternatively be interpreted as Bayes estimators. If the data-discrepancy $\LossData\bigl(\ForwardOp\x ,\y\bigr)$
is proportional to the negative log-likelihood for $(\sty | \stx=\x)$, then minimizing it corresponds to maximum-likelihood estimation. 
In addition, if the regularizer $\RegFunc_{\theta}(\x)$ in \eqref{eq:VarReg} is proportional to the negative log-prior density of $\stx$, then \eqref{eq:VarReg} can be interpreted as the \ac{MAP} estimation \cite{Pereyra2019}. 

Finally, it is important to note that the majority of physical phenomena are of nonlinear nature, but often a linearity assumption can be made under ideal measurement conditions. Nevertheless, for more advanced techniques or when accurate quantitative estimates are needed, the full nonlinear model is necessary. In this case, convergence guarantees often hold only locally \cite{kaltenbacher2008iterative}.

A relevant question in practice is now what happens if the model in the physics-driven reconstruction is inaccurately approximated or linearized. Unfortunately, in general a recovery of the true unknown can only be guaranteed if the model mismatch is taken into account \cite{arridge2006approximation}. But, as these approximation errors are often of nonlinear nature, modern data-driven approaches offer a new and promising avenue to take these into account \cite{lunz2021learned}.


\subsection{Data driven reconstruction}


Although model-based inversion such as variational regularization with highly complex and sophisticated analytical regularizers represents a promising approach to solve ill-posed inverse problems, they pose two key challenges:
\begin{inparaenum}[(i)]
\item \emph{handcrafting} a sufficiently expressive regularizer (or prior) and \item ensuring computational feasibility of hyper-parameter selection and evaluation.
\end{inparaenum}
These challenges are more pronounced for Bayesian inversion.
Algorithms to approximate the posterior mean and to perform uncertainty quantification are typically based on \ac{MCMC} methods, which require carefully constructed priors and tend to be computationally infeasible for time-critical applications. 

The development of data-driven reconstruction is inspired by the need to address the above challenges of achieving computational feasibility and selection of a domain-adapted regularizer/prior. Instead of handcrafting a reconstruction method, data-driven methods use training data to learn an \textit{optimal} reconstruction method based on statistical learning.   
We will focus here on (\emph{learned reconstruction})
methods $\RecOp_{\theta} \colon \Y \to \X$ that are typically parameterized by some suitably chosen \ac{DNN} and thus learning refers to selecting optimal parameters $\widehat{\theta}$ from the training data. This, however, depends on the statistical properties of the training data as discussed in the following and, as we will see later, has an effect on the type of convergence we obtain.

\paragraph{Supervised learning}
In this case, one has access to pairs of ground-truth images and the corresponding measurements following the measurement model $\y = \ForwardOp\x + \e$. That is, training data are given as i.i.d.\@ samples $(\x_1,\y_1)\ldots, (\x_n,\y_n) \in \X \times \Y$ of $(\stx,\sty)$. 
An optimal set of parameters $\widehat{\theta}$ for the reconstruction method is found by empirical risk minimization given a suitable loss function $\LossRec \colon \X \times \X \to \Real$ in the image domain:
\begin{equation}\label{eq:BayesEst}
\widehat{\theta} \in \argmin_{\theta} 
  \frac{1}{n}\sum_{i=1}^n
    \LossRec\bigl( \RecOp_{\theta}(\y_i), \x_i) \bigr).
\end{equation}
Usual choices for the loss function include the squared $\ell^2$-norm, where $\LossRec(\x,\x'):= \Vert \x-\x' \Vert_2^2$. So far as the training data is concerned, we assume here that we have access to appropriate training data distributions. Nevertheless, the choice of the training data plays an important role in the performance of the trained models and can lead to overly optimistic results, a phenomenon referred to as \textit{implicit data crime} by \cite{implicit_data_crime_pnas2022}.


The formulation in \eqref{eq:BayesEst} does not explicitly include the forward operator, or more generally, the data likelihood. Nevertheless, this is implicitly incorporated by the choice of training data that satisfy $\y_i \approx \ForwardOp \x_i$. One can now select a parametrization $\RecOp_{\theta} \colon \Y \to \X$ that accounts for the fact that a trained estimator should represent a regularized inversion method.
A popular example of such a domain-adapted parametrization is to combine a \ac{DNN} $\Op{C}_{\theta} \colon \X \to \X$ with a handcrafted pseudo inverse $\ForwardOp^{\dagger} \colon \Y \to \X$, which incorporates a direct regularization, as discussed in Sec. \ref{sec:modelbasedRec}. The learned reconstruction operator is then represented by the composition 
$\RecOp_{\theta} := \Op{C}_{\theta}\circ \ForwardOp^{\dagger} $ 
where $\Op{C}_{\theta} \colon \X \to \X$ acts as a learned post-processing operator that removes noise and under-sampling artifacts.
Popular architectures in imaging are based on \acp{CNN}, more specifically convolutional autoencoders with an encoding/decoding branch, such as the popular U-Net \cite{postprocessing_cnn} and related architectures inspired by harmonic analysis \cite{FrameletPaper}. The supervised setting applies to the case where pairs of high- and low-quality reconstructions are available, such as low-dose and high-dose \ac{CT} scans. In this case, the high-dose reconstruction can be identified with $x$ and the low-dose measurements provides the data $y$ to obtain an initial reconstruction by applying the pseudo-inverse operator $\ForwardOp^\dagger$, which can be considered a pre-computation step before the training procedure.

Another highly popular parameterization for $\RecOp_{\theta}$ is based on so-called \emph{unrolling}. 
The idea is to start with some iterative scheme, like one designed to minimize $\x \mapsto \LossData\bigl( \ForwardOp\x,\y\bigr)$ or the objective in \eqref{eq:VarReg}.
Next, the iterative scheme is truncated to a fixed number of iterations and unrolled by replacing selected handcrafted updates with (possibly shallow) \acp{NN}. Notably, for model-based learning, one does not usually replace $\ForwardOp$ and its adjoint $\ForwardOp^{\top}$, but other components such as the proximal operator \cite[Sec.~4.9.1]{data_driven_inv_prob}.
Hence, the \ac{DNN} for $\RecOp_{\theta}$ is formed by combining (shallow) \acp{NN} with physics-driven operators. Popular examples of unrolled networks include learned primal-dual \cite{lpd_tmi}, and variational networks \cite{hammernik2018learning}.
At this point, we would like to stress that \emph{unrolling is merely a way to select an architecture for the \ac{DNN} parametrizing  $\RecOp_{\theta}$}. In particular, training an unrolled reconstruction network following \eqref{eq:BayesEst} does not, in general, lead to an $\RecOp_{\theta}$ that minimizes a variational objective as in \eqref{eq:VarReg}, even though the architecture of $\RecOp_{\theta}$ is constructed by unrolling an optimization solver. One can instead interpret $\RecOp_{\theta}$ as a Bayes estimator that summarizes the posterior distribution of $\stx$ conditioned on $\sty=y$ using a point estimate. The statistical characterization of this estimate depends on the choice of the loss function $\LossRec$. For instance, when $\LossRec$ is the squared $\ell_2$ distance, $\RecOp_{\theta}$ approximates the classical \ac{MMSE} estimate (i.e., the conditional posterior mean $\textrm{E}\left[\stx|\sty=y\right]$) in the limit of infinite training data.

\paragraph{Unsupervised learning}
This type of learning comes in two flavors, depending on whether high-quality images or measurement data are available. In the former case, training data $\x_1,\ldots, \x_n \in \X$ are i.i.d.\@ samples of $\stx$.
One can then consider reconstruction methods $\RecOp_{\theta}$ of the form in \eqref{eq:VarReg} with a regularizer/prior $\RegFunc_{\theta} \colon \X \to \Real$ that is learned from training data through generative modeling, like \acp{GAN} or \acp{VAE}.
\Ac{GAN}-based approaches implicitly regularize the reconstruction by restricting it to the range of a pre-trained generator \cite{bora2017compressed}, whereas the other alternative is to learn an explicit regularizer parametrized by a \ac{DNN} \cite{nett_paper,ar_nips}.    

The other variant is when training data $\y_1,\ldots, \y_n \in \Y$ are i.i.d.\@ samples of $\sty$.
One option is to learn a data-driven solver for optimization problems of the type in \eqref{eq:VarReg} where the objective is parameterized by $\y \in \Y$ (see \cite{Banert:2021aa} for further details).
In this setting, it is quite natural to parametrize $\RecOp_{\theta}$ by a \ac{DNN} whose architecture is given by unrolling an optimization solver for \eqref{eq:VarReg}. Unlike the supervised setting where the reconstruction operator does not necessarily minimize a variational potential, the trained \ac{DNN} in this case is designed specifically to solve an underlying variational problem of the form \eqref{eq:VarReg}.

\paragraph{Weakly supervised learning} This refers to the case where samples of measurement data and ground-truth are available, but they are not paired. Training data then consists of un-paired $\x_1,\ldots, \x_n \in \X$ and $\y_1,\ldots, \y_n \in \Y$ in the sense that $\x_i \in \X$ and $\y_i \in \Y$ are i.i.d.\@ samples of the $\stx$- and $\sty$–marginal distributions of $(\stx,\sty)$, respectively.
Such data can be used to set-up a learning problem which quantifies consistency against data and image error similar to~\eqref{eq:BayesEst}, but using loss functions on both $\X$ and $\Y$.
This is complemented with a term that quantifies similarity of probability distributions induced by training data, thus resulting in an elaborate learning problem, see \cite[eq.~(5.3)]{data_driven_inv_prob}.
The learning problem is recast into a more suitable form by using deep generative models.

Notable methods that fall within the weakly supervised training paradigm are \ac{AR} \cite{ar_nips, uar_neurips2021} and its convex variant given by \ac{ACR} \cite{acr_arxiv}. Both of these methods seek to learn a regularizer in a variational model as a critic that can tell apart noisy reconstructions (obtained through some simple baseline approach, e.g., by applying the pseudo-inverse of $\Op{A}$ on the measurements) from the ground-truth image samples $\x_i$. The regularizer is restricted to be Lipschitz-continuous (and convex in case of \ac{ACR}), which plays an important role in the stability analysis of the resulting variational model. Lipschitz-continuity of the regularizer is enforced via a soft-penalty (see \cite{ar_nips,acr_arxiv} and references therein), whereas convexity is enforced by the choice of \ac{DNN} architecture.

\section{Types of convergence}
\label{sec:convergence_notions}
Now that we have established the concept of reconstruction methods, regularization, and learned reconstruction, the question that remains is: what are the theoretically desired properties of a learned reconstruction? For instance, if we think about iterative reconstruction, the question of convergence is relevant. Does the iterative scheme converge to a fixed point? Does this fixed point correspond to a minimizer of a variational loss, such as in \eqref{eq:VarReg}? Moreover, is it, in fact, a regularization method? These are not only purely academic questions, but provide interpretability and guarantees of correctness of the obtained solution. This section outlines important notions of convergence relevant for the image reconstruction task at hand in an accessible way, while not compromising on the mathematical rigor. 

\paragraph{Formal stability} 
The weakest and arguably the most fundamental mathematical guarantee comes in the form of \emph{stability}, which has its origin in Hadamard's definition of well-posedness \cite{benning_burger_2018}. \emph{Stability} refers to a smooth variation of the reconstruction with respect to changes in the observed data (see box).
The notion of stability can also be applied to Bayesian inversion, in which case it refers to the stability of posterior probabilities, statistical moments, and/or Bayesian estimators. When we refer to provable stability in the following, we mean that it is possible to estimate and control the maximal error in the reconstruction with respect to deviations in the measurement (for instance, in terms of the Lipschitz constant). It should be also noted, that the notion of stability (or lack thereof) itself can be quite meaningless, if no further conditions are provided. For instance, a reconstruction method that always produces the same image from varying data is in fact stable, but useless in practice.

\begin{mdframed}[style=mystyle,frametitle=Stability versus accuracy]
Consider a trained reconstruction operator $\RecOp_{\theta}$  with fixed network parameters (learned from training data). The reconstruction produced by $\RecOp_{\theta}$ is said to be \textit{stable} if $\RecOp_{\theta}:\Y \to \X$ is a continuous function of the observed data. Formally, stability demands that
\begin{equation*}
    \left\|\RecOp_{\theta}(y+w)- \RecOp_{\theta}(y)\right\|_{\X}\rightarrow 0 \text{\,\,as\,\,}\left\|w\right\|_{\Y}\rightarrow 0.
\end{equation*}
One possibility for a stability analysis is to consider the Lipschitz constant $L$ of the mapping $\RecOp_{\theta}$, which is given by the smallest $L>0$, such that
\begin{equation}
\|\RecOp_{\theta}(\y_1) - \RecOp_{\theta}(\y_2)\| \leq L \|\y_1 - \y_2\|, \text{\,\,for all\,\,}\y_1,\y_2\in \Y.
\label{eq:def_lipschitz}
\end{equation}
It is important to note that since \acp{DNN} are compositions of affine functions and smoothly varying nonlinear activations, a mapping $\RecOp_{\theta}$ modeled using \acp{DNN} is continuous and a constant $L$ satisfying \eqref{eq:def_lipschitz} exists. However, although $\RecOp_{\theta}$ is formally stable, the constant $L$ might be large, leading to large deviations in the reconstruction for small changes in the measurement.\\
Additionally, a consequence of \eqref{eq:def_lipschitz} is that the reconstruction of a slightly perturbed image must satisfy
\begin{equation*}
\|\RecOp_{\theta}(\Op{A}(x+\eta)) - \RecOp_{\theta}(\Op{A}x)\| \leq L \|\Op{A}\eta\|, \text{\,\,for any perturbation\,\,}\eta.
\label{eq:def_lipschitz_perturb}
\end{equation*}
Since the forward operator $\Op{A}$ is ill-posed and with possibly non-trivial null space, $\|\Op{A}\eta\|$ could be arbitrarily small for some perturbation $\eta$. As a result, the reconstruction remains insensitive to such changes, thereby compromising in accuracy if $L$ is small, and consequently an accurate $\RecOp_\theta$  for small perturbations must have a  large Lipschitz constant $L$.


\end{mdframed}

\begin{mdframed}[style=mystyle,frametitle=Adversarial robustness]
Adversarial robustness of a trained reconstruction operator $\RecOp_{\theta}$ is measured by the largest deviation caused in the reconstruction by a small amount of noise in the data. For a given $y_0=\Op{A}x_0\in \Y$, where $x_0$ is the underlying image, and a given noise level $\epsilon_0$, this is defined formally as \cite{GenzelTPAMI}:
\begin{equation}
    \delta_{\text{adv}} = \underset{w:\|w\|\leq \epsilon_0}{\sup}\left\|\RecOp_{\theta}(y_0+w)-\RecOp_{\theta}(y_0)\right\|_{2}.
    \label{eq: adv_robustness_def}
\end{equation}
If $\delta_{\text{adv}}$ is small for small $\epsilon_0$, the reconstruction method $\RecOp_{\theta}$ is said to be adversarially robust.\\ \indent Adversarial robustness of
image recovery methods, learning-based or classical, is crucial for their safe deployment in decision-critical applications such as medical imaging. Some skepticism about adversarial stability (or lack thereof) of deep learning-based approaches has been raised in \cite{hansen_instability}. Nevertheless, subsequent work on adversarial robustness of learned methods in \cite{GenzelTPAMI} put this concern into perspective and performed a systematic comparison of data-driven methods with the classical (and provably stable) \ac{TV}-regularized solution. In a compressed sensing experiment with random Gaussian measurements, the learned methods were found to be as robust as \ac{TV} to adversarial noise, and superior to TV for statistical noise. Further, on the fastMRI knee dataset\footnote{fastMRI data set available at: https://fastmri.med.nyu.edu/}, the learned methods were shown to be even more resilient to large adversarial perturbations as compared to \ac{TV}. Very recently, \cite{adv_artifacts_alaifari_2022} considered an $\ell_{\infty}$-norm-based measure of adversarial stability, as opposed to the $\ell_2$-norm, owing to its relevance in capturing \textit{localized reconstruction artifacts}. The authors showed that neural network-based methods are more robust to $\ell_{\infty}$-based adversarial perturbations in comparison with \ac{TV}.\\
\indent The studies of adversarial robustness as discussed above pertain to supervised methods trained end-to-end, and only concern robustness to noise in the measurement. A principled empirical and theoretical analyses are needed for a quantitative understanding of the  robustness of unsupervised methods (such as \ac{AR}, \ac{ACR}, \ac{NETT}, \ac{PnP} methods, etc.) to noise and possibly to other types of distortions that go beyond noise (e.g., forward-model and/or prior mismatch). 


\end{mdframed}




\paragraph{Fixed-point convergence}
To solve a reconstruction problem iteratively starting from an initial guess, one applies an operator updating $\Op{T} \colon \X \to \X$ on the previous iterate(s): $\x_{k+1}:=\Op{T}(\x_k)$.
The operator $\Op{T}$ typically involves the forward operator, its adjoint, and the observed data. 
It is important to determine whether the iterates converge. One such notion is \emph{fixed-point convergence}, which means that $\lim_{k\to\infty} \x_k=\x_{\infty}$, where $\x_{\infty}:=\Op{T}(\x_{\infty})$ is a fixed-point of $\Op{T}$. If fixed-point convergence holds, the iterates stabilize after sufficiently many steps, which is clearly desirable for an iterative scheme. However, it does not necessarily tell anything about what kind of solutions the iteration converges to.

\paragraph{Convergence to the minimum of a variational loss (objective convergence)}
For a stronger notion of convergence of an iterative algorithm, one can consider minimizing a variational loss, similar to \eqref{eq:VarReg}. That is, we consider the loss function of the form $\Op{J}(\x):=\LossData\bigl( \ForwardOp\x,\y \bigr) + \RegFunc(\x)$ for data $\y \in \Y$. In an optimization algorithm, an initial guess is refined iteratively by exploiting, for instance, information about the gradient $\nabla\Op{J}$ at each step to compute a minimizer.
Such an iterative scheme $\x_{k+1}:=\psi_{\vartheta_k}\bigl(\x_k,\nabla \Op{J}(\x_k)\bigr)$ with an updating rule $\psi_{\vartheta_k} \colon \X \times \X \to \X$ and iteration-dependent parameters $\vartheta:=\{\vartheta_1,\vartheta_2,\ldots\}$ is said to converge to a minimizer if $\x_{k}\to \argmin_{x\in\X}\Op{J}(x)$ as $k\to\infty$.
That is, we can now characterize the point of convergence as the minimizer of an objective function. We will refer to this notion of convergence as \emph{objective convergence} in the remainder of the paper.

\paragraph{Convergent regularization}
The strongest form of convergence we discuss here considers whether a regularized solution for an ill-posed inverse problem with (linear) forward operator $\ForwardOp \colon \X \to \Y$ tends to the solution corresponding to noise-free data $\y^0 \in \Y$ as the noise level vanishes. This could be viewed as a second level of convergence, as convergence of the iterative scheme is needed.


Formally, a regularization method can be understood as a parameterized family  $\{ \RecOp_{\theta} \}_{\theta \in \Real^d}$ of reconstruction methods. Here, the parameter $\theta$ depends on the noise level $\delta>0$, where $\Vert \y^{\delta} - \y^0 \Vert \leq \delta$ with $\y^0 := \ForwardOp \x^*$ denoting noise-free data. A regularization method is convergent if there exists a parameter choice rule $\delta \mapsto \theta(\delta, \y^{\delta})$ such that reconstructions converge to a pseudo-inverse solution as noise vanishes, i.e., $ \RecOp_{\theta(\delta, \y^{\delta})}(y^\delta) \to \ForwardOp^\dagger(y^0)$ as $\delta\to 0$.



\begin{figure*}[!h]
\centering
\begin{minipage}[t]{0.3\textwidth}
  \centering
  \vspace{0pt}
  \begin{tikzpicture}[spy using outlines={
    rectangle, 
    red, 
    magnification=3.5,
    size=1.6cm, 
    connect spies}]
    \node {\includegraphics[width=1.8in]{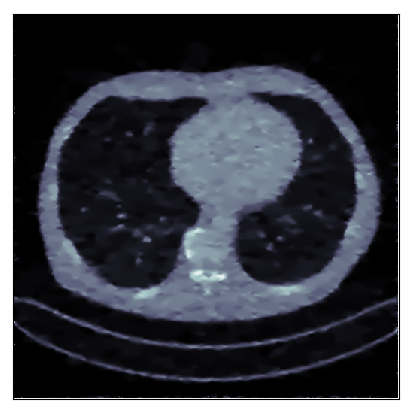}};
    \spy on (0.02,-0.85) in node [left] at (2.15,-1.35);
    \spy on (-0.60,-1.00) in node [left] at (-0.55,1.35);  
  \end{tikzpicture}
  \vskip-0.5\baselineskip
  {\small (a) $\delta=4.0$}
\end{minipage}
\hspace{0.05in}
\begin{minipage}[t]{0.3\textwidth}
  \centering
  \vspace{0pt}
  \begin{tikzpicture}[spy using outlines={
    rectangle, 
    red, 
    magnification=3.5,
    size=1.6cm, 
    connect spies}]
    \node {\includegraphics[width=1.8in]{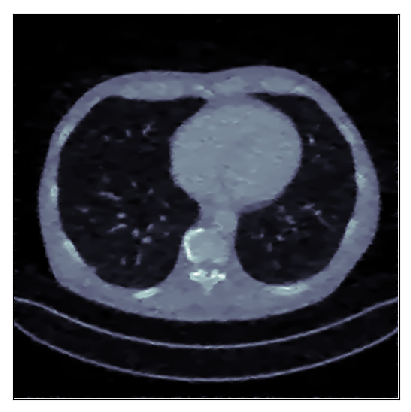}};
    \spy on (0.02,-0.85) in node [left] at (2.15,-1.35);
    \spy on (-0.60,-1.00) in node [left] at (-0.55,1.35);
  \end{tikzpicture}  
 \vskip-0.5\baselineskip
  {\small (b) $\delta=2.0$}
\end{minipage} 
\hspace{0.05in}
\begin{minipage}[t]{0.3\textwidth}
  \centering
  \vspace{0pt}
  \begin{tikzpicture}[spy using outlines={
    rectangle, 
    red, 
    magnification=3.5,
    size=1.6cm, 
    connect spies}]
    \node {\includegraphics[width=1.8in]{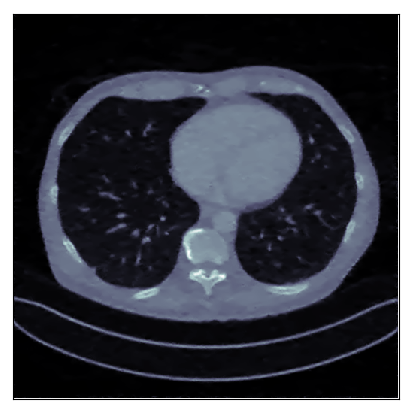}};
    \spy on (0.02,-0.85) in node [left] at (2.15,-1.35);
    \spy on (-0.60,-1.00) in node [left] at (-0.55,1.35);
  \end{tikzpicture}    
 \vskip-0.5\baselineskip
  {\small (c) $\delta=1.0$}
\end{minipage}  
\\ 
\begin{minipage}[t]{0.3\textwidth}
  \centering
  \vspace{0pt}
  \begin{tikzpicture}[spy using outlines={
    rectangle, 
    red, 
    magnification=3.5,
    size=1.6cm, 
    connect spies}]
    \node {\includegraphics[width=1.8in]{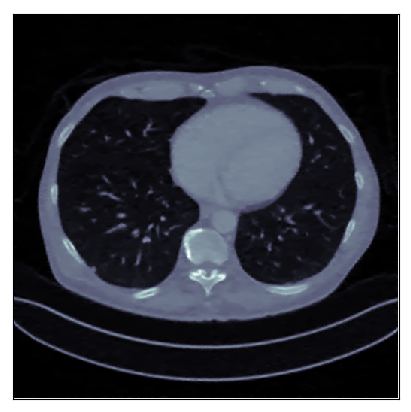}};
    \spy on (0.02,-0.85) in node [left] at (2.15,-1.35);
    \spy on (-0.60,-1.00) in node [left] at (-0.55,1.35);
  \end{tikzpicture}
  \vskip-0.5\baselineskip
  {\small (d) $\delta=0.50$}
\end{minipage}
\hspace{0.05in}
\begin{minipage}[t]{0.3\textwidth}
  \centering
  \vspace{0pt}
  \begin{tikzpicture}[spy using outlines={
    rectangle, 
    red, 
    magnification=3.5,
    size=1.6cm, 
    connect spies}]
    \node {\includegraphics[width=1.8in]{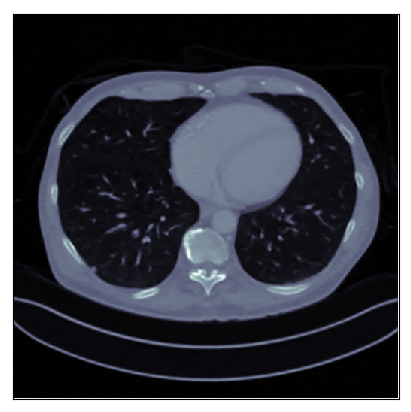}};    
    \spy on (0.02,-0.85) in node [left] at (2.15,-1.35);
    \spy on (-0.60,-1.00) in node [left] at (-0.55,1.35);
  \end{tikzpicture}
  \vskip-0.5\baselineskip  
  {\small (e) $\delta=0.25$}
\end{minipage}
\hspace{0.05in}
\begin{minipage}[t]{0.3\textwidth}
  \centering
  \vspace{0pt}
  \begin{tikzpicture}[spy using outlines={
    rectangle, 
    red, 
    magnification=3.5,
    size=1.6cm, 
    connect spies}]
    \node {\includegraphics[width=1.8in]{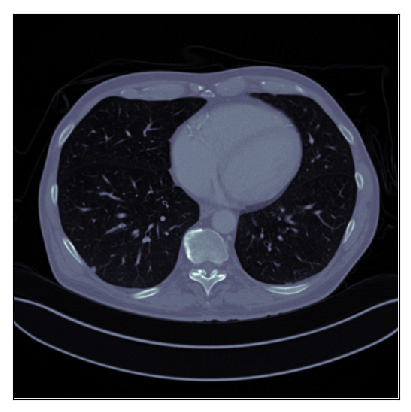}};
    \spy on (0.02,-0.85) in node [left] at (2.15,-1.35);
    \spy on (-0.60,-1.00) in node [left] at (-0.55,1.35);
  \end{tikzpicture}
  \vskip-0.5\baselineskip  
  {\small (f) Ground-truth}
\end{minipage}
\caption{\small{\ac{ACR} \cite{acr_arxiv} as a convergent regularization strategy: The reconstructed image converges to the ground-truth as $\delta\to 0$, subject to an appropriate parameter choice rule for the scalar regularization parameter: $\delta \mapsto \lambda(\delta)>0$}. Here, the regularizer parameter $\vartheta$ is learned from data and kept fixed during reconstruction. The highlighted regions show the key differences between the reconstructed images for different values of the pair $(\delta,\lambda(\delta))$.}
\label{fig:acr_convergence}
\end{figure*}

In the context of variational models \eqref{eq:VarReg}, one can re-formulate the above as follows: Let $\x_{\theta,\delta} \in \X$ denote a minimizer to the objective in \eqref{eq:VarReg} for given $\theta$ and data $\y^{\delta} \in \Y$ with noise level $\Vert \e \Vert = \Vert \y - \y^0 \Vert < \delta$.
Next, assume also that there is a parameter choice rule $\delta \mapsto \theta(\delta,\y^{\delta})$, such that $\theta(\delta,\y^{\delta})\rightarrow \theta_0$ as $\delta\rightarrow 0$.
The variational model defined by \eqref{eq:VarReg} is said to \emph{converge to an $\RegFunc$-minimizing solution} if $\x_{\theta(\delta,\y^{\delta}),\delta} \to \x^{\dagger}$ as $\delta \to 0$. Here, $\x^{\dagger} \in \X$ denotes a minimizer of the regularization functional $\RegFunc_{\theta_0}$ among all solutions that are consistent with the clean measured data $\y^0$. The $\RegFunc$-minimizing solution is formally defined as
\begin{equation}
{\x}^{\dagger}\in\argmin_{\x \in \X}\,\,
 \RegFunc_{\theta_0}(\x) 
 \quad\text{subject to $\y^0=\ForwardOp \x$,
 where $\theta_0 := \lim_{\delta \to 0}\theta(\delta,\y^{\delta})$.}
\label{R_min_def}
\end{equation}
Fig.~\ref{fig:acr_convergence} shows such convergence in the context of \ac{ACR} \cite{acr_arxiv}, a learned convex regularizer. Here, the regularizer is constructed as $\RegFunc_{\theta}(x)=\lambda\,\mathscr{S}_{\vartheta}(x)$, where the functional $\mathscr{S}_{\vartheta}$ is modeled using an \ac{ICNN} and the parameters $\vartheta$ are learned from training data. For reconstruction, the variational optimization problem in \eqref{eq:VarReg} is solved with $\theta=(\lambda,\vartheta^*)$, where $\vartheta^*$ denotes the parameter of a trained model and $\lambda:\delta\mapsto \lambda(\delta)$, the regularization penalty, should be chosen appropriately based on $\delta$ to guarantee convergence to the $\mathscr{S}_{\vartheta^*}$-minimizing solution \cite{acr_arxiv}.

\section{Provable stability and convergence}\label{sec:provable}
\begin{figure}[ht]
\centering
\includegraphics[width=0.42\linewidth]{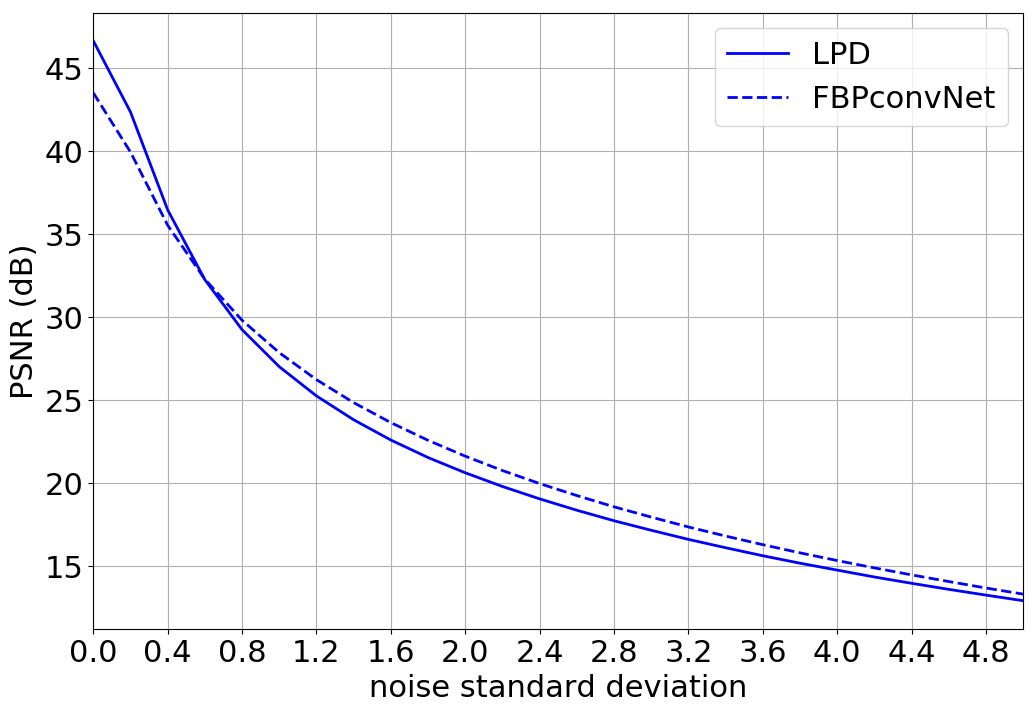}
\includegraphics[width=0.42\linewidth]{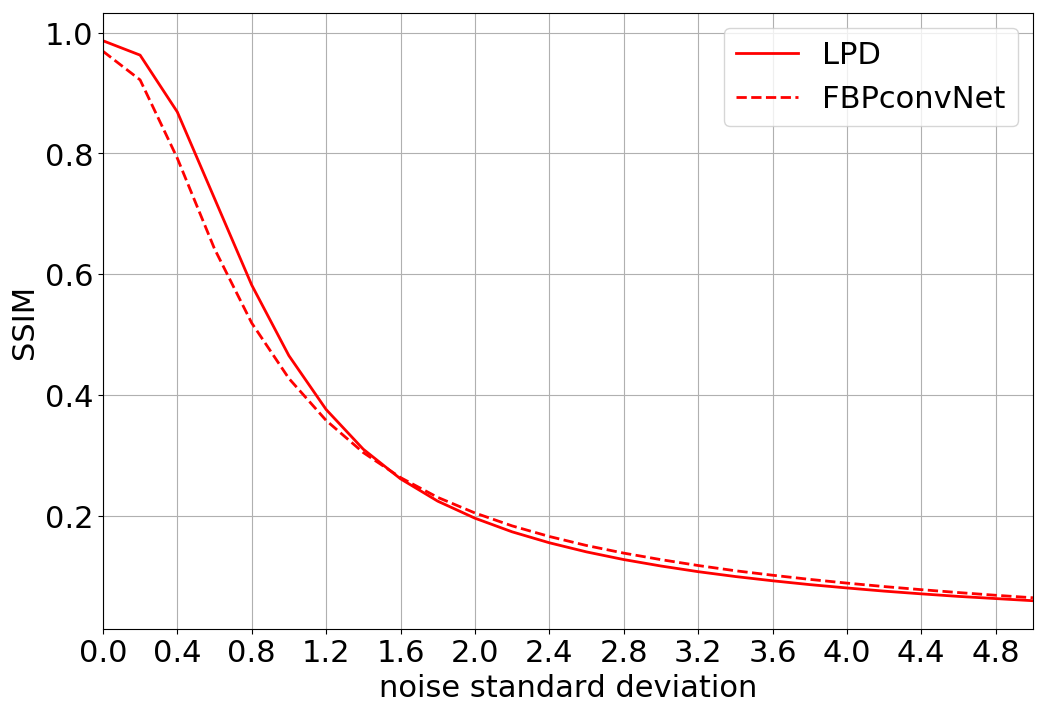}
\caption{Reconstruction quality (\acs{PSNR} (dB) on the left and \acs{SSIM} on the right) as a function of noise level in data for \ac{LPD} (unrolling) \cite{lpd_tmi} and FBPconvNet (one-step) \cite{postprocessing_cnn} methods. 
Both methods are trained against simulated noise-free 2D sparse-view \ac{CT} data generated from 2D images that are cross sections of anthropomorphic phantoms in the AAPM low-dose \acs{CT} challenge \cite{mayo_ct_challenge}.}
\label{fig:stability_wrt_noise}
\end{figure}

Most learned reconstruction methods $\RecOp_{\theta} \colon \Y \to \X$ are formally stable since they are continuous mappings.
This is the case with one-step methods $\RecOp_{\theta} := \Op{C}_{\theta} \circ \ForwardOp^{\dagger}$ whenever the pseudo-inverse $\ForwardOp^{\dagger} \colon \Y \to \X$ and the learned post-processor $\Op{C}_{\theta} \colon \X \to \X$ are continuous.
Likewise, common unrolling architectures for $\RecOp_{\theta}$, like variational networks \cite{hammernik2018learning} and \ac{LPD} \cite{lpd_tmi}, are continuous.
The above claims are supported by Fig.~\ref{fig:stability_wrt_noise}, which shows performance of FBPconvNet (one-step method) and \ac{LPD} (unrolling architecture) for sparse-view CT reconstruction.
In contrast, a reconstruction operator given by a variational scheme with a learned regularizer, as in \ac{AR}, is not necessarily continuous.
Further, even if the reconstruction operator is formally continuous, its Lipschitz constant can be large, resulting in loose stability bounds. 

\begin{figure*}[!tbh]
\centering
    \includegraphics[width=1\textwidth]{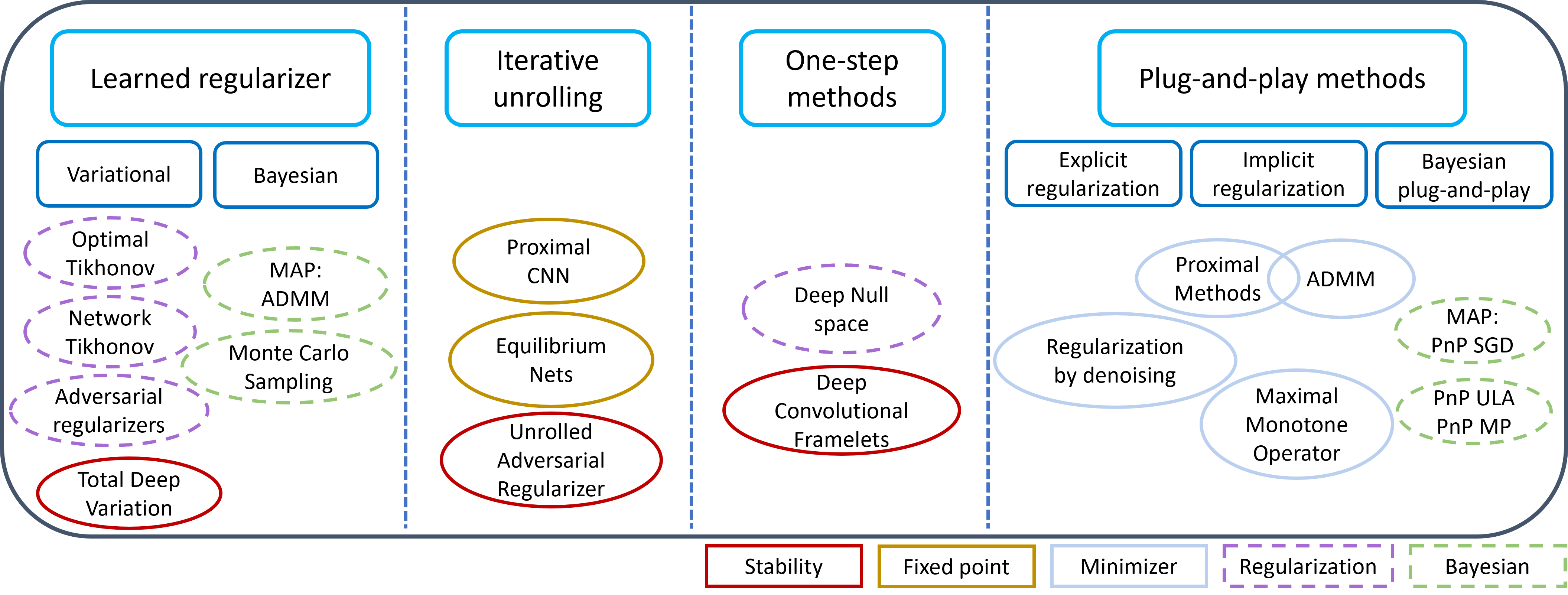}
    \caption{Categorization of data-driven reconstruction approaches, which are color-coded based on the strongest type of convergence guarantee they satisfy. 
    }
    \label{fig:Overview}
\end{figure*}

Regarding convergence, most results can be placed within the variational framework with either an explicit or an implicit regularizer, see  Fig.~\ref{fig:Overview} for an overview of the discussed methods. Explicit regularization schemes model the regularizer directly through a neural network, whereas the implicit schemes regularize the solution via a denoiser (also known as plug-and-play methods). The explicit regularization schemes \cite{ar_nips,nett_paper,acr_arxiv} typically come with a stability or convergent regularization guarantees, whereas the plug-and-play methods have been shown to possess either fixed-point or objective convergence subject to different constraints on the denoiser \cite{pnp_admm_chan_2017,gs_denoiser_hurault_2021}. 


\subsection{Learned regularization methods}
These methods are based on learning a \ac{DNN} representing a regularizer $\RegFunc_{\theta} \colon \X \to \Real$ in a variational model of the form \eqref{eq:VarReg} (learned regularizer) or reconstruction by considering images generated by a \ac{DNN} (regularization-by-architecture).

Inspired by the optimal transport theory, an adversarial framework for learning the regularizer was proposed in \cite{ar_nips}. 
This method, also referred to as the \acf{AR} method, comes with stability guarantees subject to the regularizer being 1-Lipschitz and coercive.
Its convex counterpart (abbreviated as \ac{ACR}) models the regularizer using an \ac{ICNN} \cite{amos2017input}. 
The analysis of \ac{ACR} \cite{acr_arxiv} follows from the classical convex regularization theory and one can formally establish well-posedness of the reconstruction problem (i.e., existence and uniqueness of the solution, and its continuous dependence on the observed data) using (strong) convexity of the regularizer. 
Moreover, \ac{ACR} can be shown to be a convergent regularization technique using function-analytic tools in classical regularization theory for inverse problems. 
Another approach to learning a regularizer is the \ac{NETT} method \cite{nett_paper}, which considers a learned regularizer given by a \ac{CNN} that is trained using an encoder-decoder set-up. 
The resulting variational model is shown to be well-posed and convergent subject to mild conditions (see \cite[Condition~2.2]{nett_paper}) on the neural network that parameterizes the regularizer. Other variants and extensions of \ac{NETT} (such as augmented \ac{NETT} \cite{augmented_nett_Obmann_2021} and the synthesis counterpart of \ac{NETT} \cite{synthesis_nett_Obmann_2020}) are also provably convergent regularization methods.


A slightly different approach is \cite{habring2021generative}, which can be seen as a regularization-by-architecture scheme akin to deep image prior \cite{ulyanov2018deepImagePrior}. Regularization properties for this method are shown rigorously by constructing the generator as a multi-level sparse coding network. 
The approach proposed in \cite{hegde_untrained_network_priors} is similar in spirit with \cite{habring2021generative}, in the sense that regularization is achieved by restricting the image to lie in the range of an untrained generator network. 
The paper provides recovery guarantees that are similar to the compressed sensing guarantees in flavor (using the so-called set-restricted eigenvalue condition). 
This condition is essentially the same as the \textit{restricted isometry condition} \cite[Definition 2]{candes_cs_spm}, but defined for all images in the range of the generator. 
In contrast with \cite{ulyanov2018deepImagePrior,hegde_untrained_network_priors}, the method developed in \cite{bresler_gan_projection} seeks a reconstruction in the range of a pre-trained generator. 
A \ac{PGD} algorithm is used for recovery by replacing the projection operator onto the range of the generator with a learned network. The recovery algorithm is provably convergent.  

\begin{figure*}[!h]
\centering
\begin{minipage}[t]{0.3\textwidth}
  \centering
  \vspace{0pt}
  \begin{tikzpicture}[spy using outlines={
    rectangle, 
    red, 
    magnification=3.5,
    size=1.6cm, 
    connect spies}]
    \node {\includegraphics[width=1.8in]{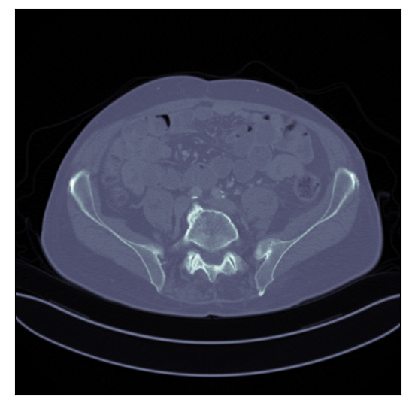}};
    \spy on (0.10,-0.68) in node [left] at (2.20,-1.30);
    \spy on (-0.54,1.00) in node [left] at (2.20,1.38);  
  \end{tikzpicture}
  \vskip-0.5\baselineskip
  {\small (a) Ground-truth}
\end{minipage}
\hspace{0.05in}
\begin{minipage}[t]{0.3\textwidth}
  \centering
  \vspace{0pt}
  \begin{tikzpicture}[spy using outlines={
    rectangle, 
    red, 
    magnification=3.5,
    size=1.6cm, 
    connect spies}]
    \node {\includegraphics[width=1.8in]{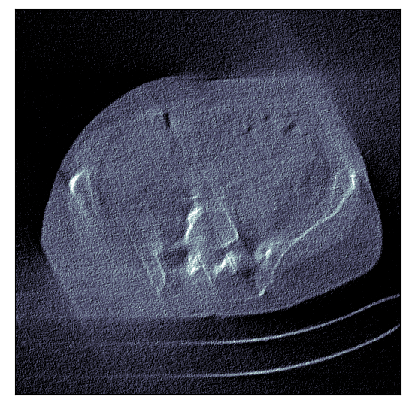}};
    \spy on (0.10,-0.68) in node [left] at (2.20,-1.30);
    \spy on (-0.54,1.00) in node [left] at (2.20,1.38);  
  \end{tikzpicture}  
 \vskip-0.5\baselineskip
  {\small (b) \Acs{FBP}: 21.61 dB, 0.17}
\end{minipage} 
\hspace{0.05in}
\begin{minipage}[t]{0.3\textwidth}
  \centering
  \vspace{0pt}
  \begin{tikzpicture}[spy using outlines={
    rectangle, 
    red, 
    magnification=3.5,
    size=1.6cm, 
    connect spies}]
    \node {\includegraphics[width=1.8in]{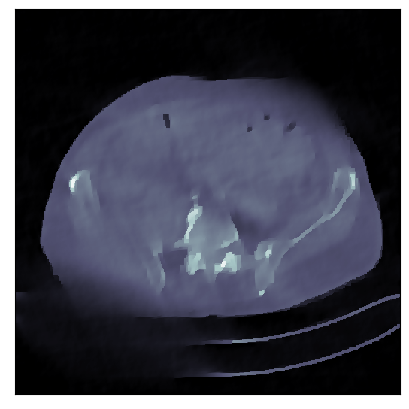}};
    \spy on (0.10,-0.68) in node [left] at (2.20,-1.30);
    \spy on (-0.54,1.00) in node [left] at (2.20,1.38);  
  \end{tikzpicture}    
 \vskip-0.5\baselineskip
  {\small (c) \Acs{TV}: 25.74 dB, 0.80}
\end{minipage}  
\\ 
\begin{minipage}[t]{0.3\textwidth}
  \centering
  \vspace{0pt}
  \begin{tikzpicture}[spy using outlines={
    rectangle, 
    red, 
    magnification=3.5,
    size=1.6cm, 
    connect spies}]
    \node {\includegraphics[width=1.8in]{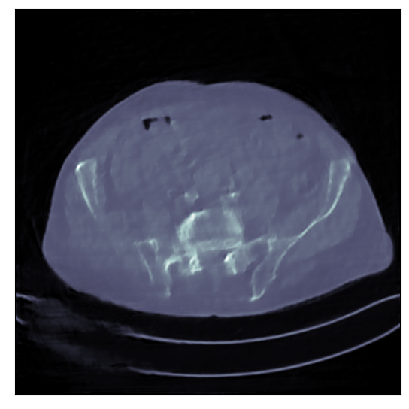}};
    \spy on (0.10,-0.68) in node [left] at (2.20,-1.30);
    \spy on (-0.54,1.00) in node [left] at (2.20,1.38);  
  \end{tikzpicture}
  \vskip-0.5\baselineskip
  {\small (d) \Acs{LPD}: 29.51 dB, 0.85}
\end{minipage}
\hspace{0.05in}
\begin{minipage}[t]{0.3\textwidth}
  \centering
  \vspace{0pt}
  \begin{tikzpicture}[spy using outlines={
    rectangle, 
    red, 
    magnification=3.5,
    size=1.6cm, 
    connect spies}]
    \node {\includegraphics[width=1.8in]{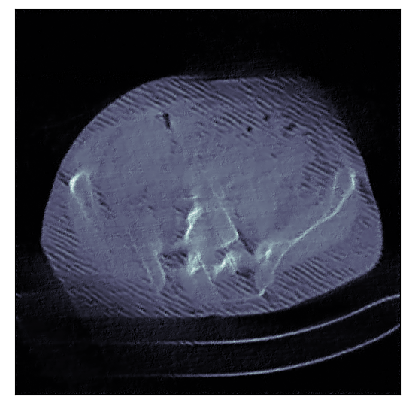}};
    \spy on (0.10,-0.68) in node [left] at (2.20,-1.30);
    \spy on (-0.54,1.00) in node [left] at (2.20,1.38);  
  \end{tikzpicture}
  \vskip-0.5\baselineskip  
  {\small (e) \Acs{AR}: 26.83 dB, 0.71}
\end{minipage}
\hspace{0.05in}
\begin{minipage}[t]{0.3\textwidth}
  \centering
  \vspace{0pt}
  \begin{tikzpicture}[spy using outlines={
    rectangle, 
    red, 
    magnification=3.5,
    size=1.6cm, 
    connect spies}]
    \node {\includegraphics[width=1.8in]{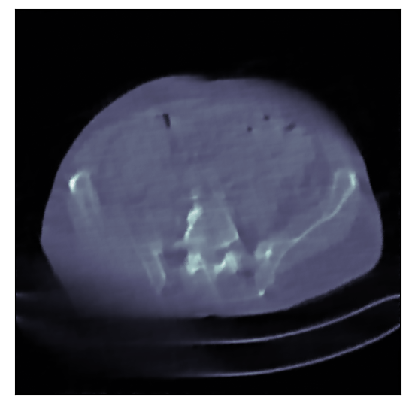}};
    \spy on (0.10,-0.68) in node [left] at (2.20,-1.30);
    \spy on (-0.54,1.00) in node [left] at (2.20,1.38);  
  \end{tikzpicture}
  \vskip-0.5\baselineskip  
  {\small (f) \Acs{ACR}: 27.98 dB, 0.84}
\end{minipage}
\caption{\small{A comparison of different model-driven and learned methods for limited-angle \ac{CT} (with the PSNR (dB) and SSIM scores indicated below). While the \ac{FBP} reconstruction is noisy, \ac{TV} fails to preserve important details in the reconstructed image. The \ac{LPD} method, which was trained on pairs of ground-truth images and limited-angle projection data in this case, does not faithfully reconstruct the image, especially the highlighted regions. One possible reason behind this is that LPD, or any other end-to-end supervised method for that matter, does not necessarily produce a data-consistent reconstruction (see Sec. \ref{sec:null_space_methods} for further explanation). The \ac{AR} leads to artifacts in the reconstructed image, as it  favors oscillations in the direction of blurring artifacts in the FBP images. Imposing convexity on the regularizer helps prevent such oscillations and resulting instability, as seen in the images reconstructed by \ac{ACR}. Notably, unlike \ac{LPD} and \ac{AR}, \ac{ACR} is a convergent regularization scheme and the reconstructions exemplify the importance of this type of theoretical guarantee. }
}
\label{lim_ct_image_figure}
\end{figure*}

\begin{mdframed}[style=mystyle,frametitle=Adversarial regularizers: Why convexity matters]
Consider a denoising problem on the real line\footnote{Thanks to Sebastian Lunz for providing this example} (i.e., $\X=\mathbb{R}$), where the distribution of ground-truth is given by $p^{\star}(x)=\frac12\bigl(\delta_{-1}(x) + \delta_{1}(x)\bigr)$, two Dirac pulses at $+1$ and $-1$. Let the noisy data have distribution $p_{\text{noisy}}=U\bigl([-\frac12,\frac12]\bigr)$, the uniform distribution over $[-\frac12,\frac12]$. Recall the adversarial learning framework \cite{ar_nips}, wherein the regularizer is trained to discern the distribution of the ground-truth from that of some baseline reconstruction by minimizing the functional
\begin{equation*}
    \mathcal{L}_{\text{AR}}(\mathscr{S})=\textrm{E}_{\stx\sim p^{\star}} \mathscr{S}(\stx) - \textrm{E}_{\stx\sim p_{\text{noisy}}} \mathscr{S}(\stx),
    \label{eq:ARTrainingLoss}
\end{equation*}
over $\mathscr{S} \in \operatorname{Lip}(\X)$, where $\operatorname{Lip}(\X)$ denotes the class of 1-Lipschitz functionals on $\X$.


\noindent Thanks to the Kantorovich-Rubinstein duality, the optimal regularizer $\mathscr{S}^*$ satisfies $\mathcal{L}_{\text{AR}}(\mathscr{S}^*)=-\mathbb{W}_1\left(p_{\text{noisy}},p^{\star}\right)$, the negative Wasserstein-1 distance between $p^{\star}$ and $p_{\text{noisy}}$. As apparent in this case, the optimal transport map from $p_{\text{noisy}}$ to $p^{\star}$ is given by
\begin{equation}
T(x)=\begin{cases}
1 & \text{ for $x\geq 0$} \\
-1 & \text{ for $x< 0$.}
\end{cases}
\label{eq:OTmap}
\end{equation}
Consequently, the Wasserstein-1 distance evaluates to:
\begin{equation*}
    \mathbb{W}_1\left(p_{\text{noisy}},p^{\star}\right)=\int_{-\frac{1}{2}}^{0}\left|x-(-1)\right|\mathrm{d}x + \int_{0}^{\frac{1}{2}}\left|x-1\right|\mathrm{d}x=\frac{3}{4}.
\end{equation*}
The function $\mathscr{S}^*(\x)=-|\x|$ is 1-Lipschitz and achieves $\mathcal{L}_{\text{AR}}(\mathscr{S}^*)=-\frac{3}{4}$, and is therefore the optimal regularizer. The resulting variational problem for denoising reads
\begin{equation}
 \min_{\x}\,\frac{1}{2}(\x-\y)^2-\lambda\,|\x|,
 \quad\text{where}\quad
 -\frac{1}{2}\leq y \leq \frac{1}{2} \text{\,\,and\,\,}\lambda>0.
 \label{eq:denoising_ex1d}
\end{equation}
One can solve \eqref{eq:denoising_ex1d} in closed form as
\begin{equation}
\hat{x}(y)=\begin{cases}
y+\lambda & \text{ for $y\geq 0$} \\
y-\lambda & \text{ for $y< 0$.}
\end{cases}
\label{eq:denoising_ex1d2}
\end{equation}
Clearly, the reconstruction given by \eqref{eq:denoising_ex1d2} changes drastically as the data $\y$ changes sign and is therefore discontinuous at $\y=0$. Note that it does not violate the stability guarantee of \ac{AR}, which only ensures convergence up to sub-sequences. For instance, consider a sequence $y_k=(-1)^k\frac{1}{k}$ for $k=1,2,\ldots$, so $y_k\to 0$ as $k\to \infty$. The corresponding sequence of reconstructions is given by $\bigl\{(-1)^k\frac{1}{k}+(-1)^k\lambda\bigr\}_{k\geq 1}$, which does not converge, but has a sub-sequence converging to $\lambda$, which is a solution of \eqref{eq:denoising_ex1d} for $y=0$.  

Imposing (strong) convexity on the regularizer \cite{acr_arxiv} helps achieve stronger forms of convergence and precludes such discontinuities in the reconstruction. More precisely, for two measurement vectors $\y_1$ and $\y_2$ that are $\delta$ apart (in norm on $\Y$), the corresponding reconstructions can vary (with respect to norm on $\X$) by at most $\frac{\beta \delta}{\lambda \rho}$, where $\beta$ is the spectral norm of $\Op{A}$ and $\mathscr{S}_{\vartheta}$ is $\rho$-strongly convex \cite[Prop. 2]{acr_arxiv}. Notably, \ac{ACR} is also a convergent regularization scheme, meaning that the reconstruction converges to the $\mathscr{S}_{\vartheta}$-minimizing solution of $\Op{A}\x=\y^{0}$, where $\y^{0}$ denotes clean data, as the noise level $\delta\rightarrow 0$, provided that the regularization penalty $\delta \mapsto \lambda(\delta)$ satisfies $\underset{\delta\rightarrow 0}{\lim}\,\lambda(\delta)=\underset{\delta\rightarrow 0}{\lim}\,\frac{\delta}{\lambda(\delta)}=0$ \cite[Prop. 3]{acr_arxiv}. The importance of convexity prior for stability is demonstrated through the example of limited-view \ac{CT} reconstruction from \cite{acr_arxiv}\footnote{Thanks in particular to Zakhar Shumaylov for the limited-view \ac{CT} experiments.} in Fig. \ref{lim_ct_image_figure}.

Notably, such convergence results for strongly convex regularizers are not limited to only linear forward operators. If the forward operator $\Op{A}$ is nonlinear, it needs to satisfy some additional technical conditions for the variational reconstruction to converge to the $\RegFunc$-minimizing solution as $\delta\rightarrow 0$. One such condition is that the level sets $L_t:=\left\{x\in \X:\Op{J}_{\theta}(\x,\y)\leq t\right\}\subset \X$ of the variational objective are required to be sequentially pre-compact for any $t>0$, meaning that, every sequence in $L_t$ must have a sub-sequence converging to some element in $\X$ (which does not necessarily have to be in $L_t$). For a complete and precise statement of such strong convergence results, we refer interested readers to Proposition 3.32 in \cite{scherzer2009variational}. 

\end{mdframed}

\subsection{Iterative unrolling with fixed-point convergence}
The main philosophy behind algorithm unrolling is to construct a \ac{DNN} architecture by unfolding a fixed number of iterations of an optimization algorithm. Subsequently, different components of the algorithm are replaced by learnable units (typically modeled using shallow neural networks) and the overall network is trained end-to-end to produce a reconstruction from its corresponding measurement. The origin of unrolling can be traced back to the seminal work by Gregor and LeCun \cite{lecun_ista} for solving sparse coding via unfolding the iterative soft-thresholding algorithm. The output of the $k^{\text{th}}$ layer of a generic unrolled architecture can be expressed as $x_{k+1}=\psi_{\theta_k}(\x_k,\y)$, where $\psi_{\theta_k}$ is a non-linear mapping with parameters $\theta_k$. 

As we remarked in Sec. II.B.a, the reconstruction of an end-to-end trained unrolled network can be interpreted as a Bayes estimator. Recall that the conditional mean estimator, given by $\Op{R}^*(y)=\textrm{E}\left[\stx|\sty=y\right]$ is a solution of
\begin{equation}
    \Op{R}^* = \underset{\Op{R}:\Y\rightarrow \X}{\argmin}\,\textrm{E}_{\stx,\sty}\left\|\Op{R}(\sty)-\stx\right\|_{2}^2,
\end{equation}
where the minimization is carried out over all measurable mappings from $\Y$ to $\X$ (see Proposition 2 in \cite{deep_bayes_inv}). Therefore, an unrolled network with a sufficiently powerful parameterization to approximate any measurable map from the data space to the image space essentially seeks to approximate $\Op{R}^*$ given enough training data. One can obtain an approximation to a different Bayes estimator by using a different loss function for training.

However, without any further assumptions on $\psi_{\theta_k}$, it is generally not possible to characterize the estimate of an unrolled network as the stationary point of a variational potential or a fixed-point of a non-linear map. Further, one trains an unrolled architecture for a few iterations (typically $\leq 20$) due to memory constraints, and the reconstruction deteriorates if more iterations are performed at test time than the number of iterations that were used in training. 

The deep equilibrium (DEQ) model proposed in \cite{deq_bai_neurips2019} provided a promising avenue to reduce the memory requirement of training unrolled networks. The key idea was to represent the output of a feed-forward model as a fixed-point of a nonlinear transformation, through which one can back-propagate using implicit differentiation. This approach can effectively learn an infinite depth network with a constant memory footprint. The DEQ models were leveraged in \cite{gilton2021deep_eq_models} for imaging inverse problems. Such models come with two notable characteristics: (i) weight-sharing, i.e., by using the same set of parameters at each layer ($\theta_k=\theta$ for all $k$), and (ii) explicitly constraining the output $\x^*(\y,\theta)$ of the unrolled network to be a fixed-point of $\psi_{\theta}$, while further ensuring that the (nonlinear) operator $\psi_{\theta}$ is contractive. The training problem reads 
\begin{equation}
    \underset{\theta}{\min}\,\frac{1}{n}\sum_{i=1}^{n}\left\|\x_i^*(\y_i,\theta)-\x_i\right\|_2^2\text{\,\,subject to\,\,}\x_i^*(\y_i,\theta) = \psi_{\theta}\left(\x_i^*(\y_i,\theta),\y_i\right).
    \label{eq:dem_training}
\end{equation}
It is shown in \cite{gilton2021deep_eq_models} that the trained map $\psi_{\theta}$ can be applied iteratively beyond the number of iterations trained without any degradation in reconstruction quality. The DEQ architectures come with fixed-point convergence incorporated into them by construction.    

Back-propagating through an infinite-depth network in DEQ models essentially requires computing inverse Jacobian-vector product, which is approximated using fixed-point iterations \cite[Sec. 4.2]{gilton2021deep_eq_models} or via quasi-Newton (QN) methods, namely Broyden methods \cite[Sec. 3.1.3]{deq_bai_neurips2019}. Such an iterative approximation of inverse Jacobian-vector products could be computationally demanding, which slows down the training of DEQ models. Some recent works have sought to accelerate the training of DEQ networks, for instance via Jacobian-free back-propagation \cite{jfb_fung}, or by using the SHINE approach in \cite{ramzi2022shine}, which considered QN methods to compute the forward pass of DEQ models and used the QN matrices (that are available as a bi-product of the forward pass) to approximate the inverse Jacobian.

Generally, sharing the weights of different layers in an unrolling scheme leads to a less expressive model and the image quality might suffer as a consequence. However, we would like to emphasize that weight-sharing is not essential for having a provable unrolled scheme. For instance, the stochastic unrolled network proposed in \cite{lspd_tang_2021} is a memory-efficient unrolling scheme that comes with recovery guarantees, but with no weight-sharing across the layers. Learned optimization solvers (see Sec. \ref{sec:learned_opt_solvers} and \cite{Banert:2021aa} for more details) also rely on unrolling, but without any weight-sharing, and can be shown to converge to the minimizer of a variational potential.

Notably, unrolling an iterative scheme can be constructed to have objective convergence. The key idea behind such constructions is to parametrize the neural network in a manner such that it corresponds to the proximal operator of an underlying proper, convex, and lower semi-continuous function.
One example is Parseval proximal neural network \cite{ppnn_arxiv} that use tight frames to parametrize the affine layers. 

\subsection{One-step methods}
\label{sec:null_space_methods}
The one-step methods consist in learning a deep neural network-based post-processing of a model-based reconstruction based on pairs of input and target images \cite{postprocessing_cnn}. More specifically, the reconstruction operator is parameterized as $\RecOp_{\theta} := \Op{C}_{\theta}\circ \Op{B}$, where $\Op{B} \colon \Y \to \X$ denotes a classical reconstruction method (with no or few tune-able parameters, e.g., \ac{FBP} or \ac{TV} in X-ray \ac{CT}) and $\Op{C}_{\theta} \colon \X\to \X$ represents a deep convolutional network with parameters $\theta$. The reconstructed image produced by such a post-processing method fails to satisfy the data-consistency criterion. That is, a small value of $\left\|\ForwardOp \x^{\dagger}- \y^{\delta}\right\|$ does not necessarily imply a small value for the data-fidelity term $\left\|\ForwardOp \Op{C}_{\theta}(\x^{\dagger})- \y^{\delta}\right\|$ corresponding to the output of $\Op{C}_{\theta}$, where $\x^{\dagger}$ is the reconstruction obtained using $\Op{B}$. Consequently, such post-processing schemes do not lead to convergent regularization strategies. This issue was addressed in \cite{Schwab_2019_null_space} by parametrizing the operator $\Op{C}_{\theta}$ as $\Op{C}_{\theta} = \text{Id}+\left(\text{Id}-\ForwardOp^{\dagger}\ForwardOp\right)\Op{Q}_{\theta}$, where $\Op{Q}_{\theta}$ is a Lipschitz-continuous \ac{DNN}. Since $\left(\text{Id}-\ForwardOp^{\dagger}\ForwardOp\right)$ is the projection operator onto the null-space of $\ForwardOp$, the operator $\Op{C}_{\theta}$ (referred to as null-space network) always satisfies $\ForwardOp\Op{C}_{\theta}(\x^{\dagger})=\ForwardOp \x^{\dagger}$, ensuring that the output of $\Op{C}_{\theta}$ explains the observed data. Null-space networks are shown to provide convergent regularization schemes  \cite{Schwab_2019_null_space}.

Deep convolutional framelets \cite{FrameletPaper} aim to gain a better understanding and interpretability of deep learning by establishing a link with wavelet theory. An image is here represented by convolving local and non-local bases. The convolutional framelets generalize the theory of low-rank Hankel matrix approaches for inverse problems and \cite{FrameletPaper} extends the idea to obtain a \ac{DNN} using multilayer convolutional framelets that enable perfect representation of an image.
The analysis in \cite{FrameletPaper} shows that the popular deep network components such as residual block, redundant filter channels, and concatenated \ac{ReLU} do indeed offer perfect representation, while the pooling and unpooling layers should be augmented with high-pass branches to meet the perfect representation condition.

\subsection{Plug-and-play denoising}
\label{sec:pnp}
Reconstruction in the variational setting typically requires an iterative algorithm to minimize the underlying variational loss. Popular iterative techniques such as \ac{PGD} and \ac{ADMM} entail applying the proximal operator corresponding to the (possibly non-smooth) regularizer to update the estimate. The seminal work by Venkatakrishnan et al. \cite{venkat_pnp_6737048} pioneered the idea of replacing the proximal operator with an off-the-shelf denoiser. To give a specific example, such a \ac{PnP}-denoising used inside the \ac{PGD} algorithm leads to the iterative scheme $\x_{k+1} = D_{\sigma}\left(x_k-\eta\nabla_{\x} \LossData\bigl( \ForwardOp\x,\y \bigr)\big|_{\x=\x_k} \right)$ for reconstruction, starting from an initial estimate $\x_0$. Here, $D_{\sigma} \colon \X \to \X$ is typically a denoiser to remove Gaussian noise of standard deviation $\sigma$. Traditionally, the choice of the denoiser has been model-inspired (e.g., BM3D, dictionary-based denoisers like K-SVD, \ac{TV}, etc.), but more recently, \ac{PnP} algorithms have used off-the-shelf deep neural network-based denoisers (e.g., DnCNN). These methods have shown excellent empirical performance for practical inverse problems, which inspired a recent line of research to analyze their convergence theoretically. Generally, \ac{PnP} methods, subject to appropriate conditions on the denoiser, have been shown to possess fixed-point and/or objective convergence.   


One of the first results on the global objective convergence of \ac{PnP}-\ac{ADMM} was shown in \cite{SreehariVWDSB15}. Their theorem requires that the denoiser is continuously differentiable and has a doubly-stochastic gradient matrix, which are equivalent to the denoiser being a proximal operator for some convex function. Fixed-point convergence of \ac{PnP}-\ac{ADMM} with a continuation scheme and bounded denoisers was established in \cite{pnp_admm_chan_2017}. It was shown in \cite{pmlr-v97-ryu19a} that the iterations of \ac{PnP}-\ac{PGD} and \ac{PnP}-\ac{ADMM} are contractive (and, hence converges to a fixed-point) if the denoiser satisfies a `Lipschitz-like' condition (see Assumption A in \cite{pmlr-v97-ryu19a}). Both objective and fixed-point convergence of \ac{PnP}-\ac{FBS} and \ac{PnP}-\ac{ADMM} were proved in \cite{kunal_9380942} for linear denoisers $D(\x)=\boldsymbol{W}\x$, where $\boldsymbol{W}$ is diagonalizable with eigenvalues in $[0,1]$. The denoiser scaling approach, which provides a systematic way to control the regularization of \ac{PnP} denoisers, is shown to be fixed-point convergent using the \ac{CE} framework \cite{xu2020boosting}.

Typically, the convergence results for \ac{PnP} methods either assume the denoiser to satisfy a hard Lipschitz bound, or require the data-fidelity term to be strongly-convex in $\x$. While the former is restrictive for deep denoisers, as it affects the denoising performance adversely, the latter assumption excludes inverse problems where the forward operator has a non-trivial null-space (e.g., sparse-view \ac{CT}, or compressed sensing). Recently, convergence guarantees for \ac{PnP} methods were derived in \cite{gs_denoiser_hurault_2021} with \ac{GS} denoisers, alleviating the need for such restrictive assumptions. \ac{GS} denoisers are constructed as $D_{\sigma}=\text{Id}-\nabla g_{\sigma}$, where $g_{\sigma}(\x)=\frac{1}{2}\left\| \x-\Op{P}_{\sigma}(\x)\right\|_2^2$, with $\Op{P}_{\sigma}$ being a deep network without any structural constraints. This parametrization was shown to have enough expressive power to achieve state-of-the-art denoising performance in \cite{gs_denoiser_hurault_2021}. Notably, \ac{GS} denoisers have a scalar potential given by $h_{\sigma}(\x)=\frac{1}{2}\left\|\x\right\|_2^2-g_{\sigma}(\x)$. When the potential $h_{\sigma}$ is convex, \ac{GS} denoisers are proximal operators corresponding to a potentially non-convex function. Therefore, this approach can target to minimize a variational objective with a potentially non-convex regularizer.

A closely related \ac{PnP} approach was adopted in \cite{mmo_pesquet} with a similar aim of providing an asymptotic characterization of iterative \ac{PnP} solutions. The main idea was to model \ac{MMO} using a \ac{NN} and interpret the reconstruction as the solution of a monotone inclusion problem (which generalizes convex optimization problems). The parametrization of \acp{MMO} was done by modeling the resolvent via a non-expansive \ac{NN}.  

\Ac{RED} is another prominent \ac{PnP} approach that utilizes an off-the-shelf denoiser $D(\x)$ to construct an explicit regularizer $\RegFunc(\x)=\lambda\cdot\x^{\top}\left(\x-D(\x)\right)$. If the denoiser is such that the gradient condition $\nabla \RegFunc(\x)=\lambda\cdot(\x-D(\x))$ holds, the \ac{RED} algorithms recover the stationary point of $\x\mapsto \frac{1}{2}\|\ForwardOp\x-\y\|_2^2+\RegFunc(\x)$. It was shown in \cite{red_schniter} that the gradient condition does not hold for denoisers with a non-symmetric Jacobian, which is the case for most practical (classical or data-driven). A new analysis framework based on score-matching was developed in \cite{red_schniter} to explain the empirical success of the \ac{RED} algorithms. Specifically, if the denoiser is such that $\frac{D(\x)-\x}{\epsilon}$ approximates the score function (i.e., the gradient of the log prior), the stationarity condition for the \ac{MAP} estimation problem (with the prior replaced by a smooth surrogate) leads to the following fixed-point equation (see Sec.~IV.C in \cite{red_schniter}): $ \frac{1}{2}\ForwardOp^{\top}\left(\ForwardOp\x^*-\y\right)+\lambda\,(\x^*-D(\x^*))=0, \text{\,\,where\,\,}\lambda=\frac{\sigma^2}{\epsilon}$,
with $\sigma^2$ being the variance of measurement noise. The equation above is identical to the fixed-point equation that the \ac{RED} algorithm seeks to recover. Here, the denoiser $D$ is not required to have a symmetric Jacobian. With this new interpretation, several variants of the \ac{RED} algorithm were proposed in \cite{red_schniter} with fixed-point convergence. 

\begin{mdframed}[style=mystyle,frametitle=Objective convergence of \ac{PnP} with \ac{GS} denoisers \cite{gs_denoiser_hurault_2021}]
The convergence of \ac{PnP} denoisers used with \ac{HQS} was established in \cite{gs_denoiser_hurault_2021}. The denoiser is constructed as a \ac{GS} denoiser as explained in Sec. \ref{sec:pnp}, i.e., $D_{\sigma}=\text{Id}-\nabla g_{\sigma}$, where $g_{\sigma}$ is proper, lower semi-continuous, and differentiable with an $L$-Lipschitz gradient. The \ac{PnP} algorithm proposed in \cite{gs_denoiser_hurault_2021} takes the form $\x_{k+1} = \text{prox}_{\tau \, f}\left(x_k-\tau \,\lambda \,\nabla g_{\sigma}(\x_k)\right)$, where $f \colon \Real^d\to\Real\cup \{+\infty\}$ measures the data-fidelity and is assumed to be convex and lower semi-continuous. Under these assumptions on $f$ and $g_{\sigma}$, the following guarantees hold for $\tau<\frac{1}{\lambda\, L}$:
\begin{enumerate}
    \item The sequence $F(\x_k)$, where $F=f+\lambda\, g_{\sigma}$, is non-increasing and convergent.
    \item $\left\|\x_{k+1}-\x_k\right\|_2 \to 0$, which indicates that iterations are stable, in the sense that they do not diverge if one iterates indefinitely. 
    \item All limit points of $\{\x_k\}$ are stationary points of $F(\x)$. 
\end{enumerate}
\end{mdframed}

\begin{mdframed}[style=mystyle,frametitle=Fixed-point convergence of \ac{PnP} with Douglas–Rachford splitting (PnP-DRS) \cite{pmlr-v97-ryu19a}]
Consider the PnP-DRS algorithm, given by
\begin{eqnarray}
x_{k+\frac{1}{2}}=\text{prox}_{\tau \, f}\left(z_k\right), x_{k+1} = D_{\sigma}\left(2x_{k+\frac{1}{2}}-z_k\right), \text{\,\,and\,\,} z_{k+1}=z_k+x_{k+1}-x_{k+\frac{1}{2}}.
\label{pnp_drs1}
\end{eqnarray}
Here, $f$ denotes the data-fidelity term and is assumed to be $\mu$-strongly convex. One can equivalently express \eqref{pnp_drs1} as the fixed-point iteration $z_{k+1}=\Op{T}(z_k)$, where
\begin{eqnarray}
\Op{T}=\frac{1}{2}\text{Id} + \frac{1}{2}\left(2D_{\sigma}-\text{Id}\right)\left(2\text{prox}_{\tau \, f}-\text{Id}\right).
\label{pnp_drs_fp}
\end{eqnarray}
Suppose, the denoiser satisfies
\begin{equation}
   \left\|\left(D_{\sigma}-\text{Id}\right)(u)-\left(D_{\sigma}-\text{Id}\right)(v)\right\|_2 \leq \epsilon  \left\|u-v\right\|_2, 
    \label{cond_denoiser_pnpDRS}
\end{equation}
for all $u,v\in \X$ and some $\epsilon>0$. It was shown in \cite{pmlr-v97-ryu19a} that if the strong convexity parameter $\mu$ is such that $\displaystyle\frac{\epsilon}{(1+\epsilon-2\epsilon^2)\,\mu}<\tau$ holds, the operator $\Op{T}$ is contractive and the PnP-DRS algorithm is fixed-point convergent. That is, $\left(x_k,z_k\right)\rightarrow (x_{\infty},z_{\infty})$, where $(x_{\infty},z_{\infty})$ satisfy 
\begin{eqnarray}
x_{\infty}=\text{prox}_{\tau \, f}\left(z_{\infty}\right) \text{\,\,and\,\,} x_{\infty} = D_{\sigma}\left(2x_{\infty}-z_{\infty}\right).
\label{pnp_drs_final}
\end{eqnarray}
As remarked in \cite{pmlr-v97-ryu19a}, the convergence of PnP-DRS follows from monotone operator theory if $\left(2D_{\sigma}-\text{Id}\right)$ is non-expansive, but \eqref{cond_denoiser_pnpDRS} imposes a less restrictive condition on the denoiser.
\end{mdframed}


\subsection{Learned optimization solvers}
\label{sec:learned_opt_solvers}
Reconstruction in imaging inverse problems is framed as an optimization problem as in \eqref{eq:VarReg}, which could be computationally demanding to solve, especially when the image lives in a high-dimensional vector space. Some recent works \cite{Banert:2021aa} have developed data-driven solvers with convergence guarantees for minimizing convex variational objectives. They seek to learn a solver for a family of optimization problems of the form $\underset{x\in\X}{\min}\,{F_y(x)}$, parametrized by $y$. In the context of inverse problems, the functional $F_y(x)$ is the variational objective defined in \eqref{eq:VarReg}. The key idea is to build a parametric solver of the form $\Op{T}_{N,\theta} \colon \Y\to \X$ by unrolling a fixed number of iterations (denoted as $N$) of a gradient-based algorithm. Subsequently, the parameters $\theta$ of the solver are learned in an unsupervised manner by minimizing $\frac{1}{n}\sum_{i=1}^{n}F_{y_i}\left(\Op{T}_{N,\theta}(y_i)\right)$ over $\theta$, where $(y_i)_{i=1}^{n}$ are $n$ i.i.d. samples drawn from the marginal distribution of the data $y$. The iterative solver is constructed by adding a neural network-based deviation term to the gradient-based update, and convergence is shown in the case where the deviation term lies in an appropriately defined set. In practice, such learned solvers converge significantly faster than a conventional first-order solver with suitably chosen step-size parameters. See Sections 3 and 4 in \cite{Banert:2021aa} for more technical details about the construction and convergence proof of learned optimization solvers. Similar provably convergent data-driven optimization solvers based on mirror-descent with an \ac{ICNN}-based Bregman distance were recently proposed in \cite{learned_MD_hyt}.


\section{Provable learned Bayesian methods}
\label{sec:provable_bayesian_methods}
\subsection{Bayesian inference in imaging inverse problems}
As explained in Sec. \ref{sec:math_foundation}, the Bayesian framework represents the unknown image $\x^*$ as a realization of a random variable $\stx$ taking values in $\X$ according to a prior distribution with density $p(x)$. The measured data is modeled by a $\Y$-valued random variable $\sty$ that is related to $\stx$ through the measurement equation $\sty =\ForwardOp\stx+ \ste$. The observed data $\y \in \Y$ is then understood as a realization of $(\sty|\stx = \x^*)$. The statistical representation of this forward model is given by the conditional density $p(y|x)$, which is known in the literature as the data likelihood function. Given the prior and the likelihood, one can use Bayes' theorem to derive the posterior distribution for $(\stx|\sty=y)$ with density given by \cite{RobertChoice}
\begin{equation}
    p(\x|\y) = \frac{p(y|x)p(x)}{\int_{\X} p(y|\tilde{x})p(\tilde{x})\mathrm{d}\tilde{x}}.
    \label{eq:bayes_posterior}
\end{equation}
The posterior distribution describes how the probability mass is distributed over the solution space $\X$ and plays a central role in Bayesian inversion. The posterior distribution not only underpins Bayesian estimators such as the minimum mean-squared error (MMSE) estimator, given by the posterior mean $\displaystyle\textrm{E}[\stx|\sty=\y] = \int_{\X} \x\, p(\x|\y)\,\textrm{d}\x$, but also has an important role in uncertainty quantification and model selection techniques \cite{RobertChoice}.

From a computational perspective, Bayesian inference often requires computing posterior probabilities and expectations, which is challenging because of the high-dimensional integrals involved. Bayesian computational algorithms based on stochastic sampling address this difficulty by using Monte Carlo integration, wherein one constructs a sequence of random variables $\{\tilde{\stx}_i\}_{i=1}^m$ such that averages computed along the sequence coincide with the desired probabilities and expectations as $m\rightarrow\infty$. For instance, the posterior mean is approximated by the Monte Carlo estimator $\displaystyle\textrm{E}[\stx|\sty=\y] \approx \frac{1}{m}\sum_{i=1}^m \tilde{\stx}_i$, with the approximation error vanishing as $m \rightarrow \infty$. The same strategy is used to compute posterior probabilities and other quantities of interest, e.g., higher-order statistical moments that can be used for uncertainty quantification \cite{Robert2004}. 


Constructing a sequence $\{\tilde{\stx}_i\}_{i=1}^m$ leading to provably fast converging Monte Carlo estimators is challenging when the dimension of $\X$ is large. The main approach in imaging is to construct such a sequence iteratively, as a Markov chain, by using a stochastic update rule that contracts the iterates towards $(\stx|\sty=\y)$ and has it as its unique fixed point. In philosophy, this is similar to iterative optimization schemes that contract iterates towards a fixed-point of interest, except for the important distinction that one works with random variables here. This approach is known as Markov chain Monte Carlo (MCMC) \cite{Robert2004} and plays an instrumental role in Bayesian inferencing.


Convergence results for MCMC algorithms characterize how quickly the iterates contract towards $(\stx|\sty=\y)$ w.r.t. a given probability measure distance, with different distances describing different forms of convergence. MCMC algorithms that contract geometrically fast as $m$ increases are particularly interesting for Bayesian computation. A detailed convergence analysis seeks to explicitly bound the distance to $(\stx|\sty=\y)$ as a function of the number of iterations $m$ and the initialization, and also characterizes the dependence of the geometric contraction rate on various key aspects of the problem (e.g., dimension, conditioning, and tail behavior).  In principle, such results allow bounding the number of iterations $m$ that are required to reach a desired numerical precision, but the bounds are too loose to be useful in practice. Instead, $m$ is often set by monitoring the quantities of interest and stopping the algorithm once these quantities are sufficiently stable. It is also recommended to discard the first $10\%$ of the Markov chain to reduce the initialization bias. Improving MCMC convergence theory to provide sharper bounds that are useful for setting $m$ is an active research area.


Moreover, while conventional Bayesian computation approaches are asymptotically unbiased, i.e., they converge exactly to $(\stx|\sty=\y)$, modern large-scale strategies often accept some bias, by allowing convergence to a controlled neighborhood of $(\stx|\sty=\y)$ to achieve significantly faster convergence rates. Similarly as optimization methods, the convergence properties of MCMC methods depend crucially on the regularity properties of the posterior $p(x|y)$ (e.g., smoothness, convexity, and tail behavior). Checking that $p(x|y)$ satisfies relevant regularity properties is challenging for Bayesian modeling strategies that use learned priors encoded by neural networks, which we will discuss later in this section. 

\begin{mdframed}[style=mystyle,frametitle=Distances between probability measures for studying convergence of Markov chains]
To define probability measures, one needs an underlying Borel-measurable space $(\X,\mathcal{B}(\X))$, where $\mathcal{B}(\X)$ is the Borel $\sigma$-algebra of $\X$. It is the smallest $\sigma$-algebra generated by the open subsets of $\X$ and contains all those subsets of $\X$ to which one can assign probabilities consistently using a probability measure. Any probability measure on $(\X,\mathcal{B}(\X))$ is formally defined as a function that maps $\mathcal{B}(\X)$ to the interval $[0,1]$.\\
Consider two such probability measures $\pi_1$ and $\pi_2$ defined on $(\X,\mathcal{B}(\X))$. A notion of distance between $\pi_1$ and $\pi_2$ is generally of the form 
\begin{equation}
    \mathcal{D}(\pi_1,\pi_2)=\underset{f\in\mathscr{F}}{\sup}\,\left|\int f\mathrm{d}\pi_1-\int \mathcal{Q}f\mathrm{d}\pi_2\right|,
    \label{eq:prob_dist_sm}
\end{equation}
where $\mathscr{F}$ is some class of functions defined on $\X$ and $\mathcal{Q} : \mathscr{F} \mapsto \mathscr{F}$.\\
When $\mathscr{F}=\left\{\mathbb{I}_{\mathscr{A}}; \mathscr{A}\in\mathcal{B}(\X)\right\}$ is the class of all indicator functions of the Borel subsets of $\X$ and $\mathcal{Q}f=f$, \eqref{eq:prob_dist_sm} leads to the well-known total variation (TV) distance given by
\begin{equation}
    \text{TV}(\pi_1,\pi_2)=\underset{\mathscr{A}\in \mathcal{B}(\X)}{\sup}\,\left|\pi_1(A)-\pi_2(A)\right|.
    \label{eq:tv_sm}
\end{equation}
In other words, the total-variation distance measures the largest deviation in the probabilities assigned by the two measures to any event $\mathscr{A}\in \mathcal{B}(\X)$. For Markov chain Monte Carlo methods, convergence in TV metric implies that probabilities computed from the Markov chain converge to the posterior probabilities of interest.

In the case where $\mathscr{F}$ contains all 1-Lipschitz functions on $\X$ and $\mathcal{Q}f=f$, \eqref{eq:prob_dist_sm} recovers the Wasserstein-1 distance between $\pi_1$ and $\pi_2$. Similarly, \eqref{eq:prob_dist_sm} leads to the Wasserstein-2 distance between $\pi_1$ and $\pi_2$ when $\mathscr{F}$ contains all bounded continuous functions on $\X$ and
\begin{equation*}
    (\mathcal{Q}f)(x)=\underset{u\in\mathscr{\X}}{\sup}\,\, f(u)-\|u-x\|_2^2,
\end{equation*}
for any $x\in \X$. Wasserstein distances play a complementary role to the TV distance in the analysis of MCMC methods. Unlike TV, convergence in a Wasserstein-$p$ distance guarantees the convergence of expectations of bounded continuous functions, as well as the convergence of the first $p$ statistical moments, but not convergence of probabilities.
\end{mdframed}

Convergence issues aside, theoretical analysis of a Bayesian imaging procedure might also seek to prove that the posterior distribution associated with $(\stx|\sty = y)$ is well-posed, and that the posterior quantities of interest exist and inherit this stability. A Bayesian inverse problem is said to be \emph{well-posed} if the posterior distribution of $(\stx|\sty = y)$ is well-defined and unique, and varies continuously w.r.t. $y$ under a given distance metric $\mathcal{D}$ (e.g. the total-variation (TV) or the Wasserstein-$p$ distances). Formally, we consider two data realizations $y_1 \in \Y$ and $y_2 \in \Y$ and denote by $\pi_1$ and $\pi_2$ the two probability measures related to the posterior distributions of $(\stx|\sty = y_1)$ and $(\stx|\sty = y_2)$. Stability in the Bayesian setting requires that $\mathcal{D}(\pi_1,\pi_2)$ is a continuous function w.r.t. $(y_1,y_2) \in \Y \times \Y$. Different choices for $\mathcal{D}$ can capture different forms of stability: well-posedness in the TV distance guarantees that the posterior probabilities are stable w.r.t. perturbations in $y$, whereas well-posedness in a Wasserstein-$p$ distance describes the stability of expectations of bounded continuous functions (but not probabilities), as well as the stability of the first $p$ posterior moments \cite{Latz2020}.

\subsection{Bayesian inference with learned priors}
The recent advances in data-driven modeling have inspired Bayesian strategies similarly as for variational regularization. In this case, instead of hand-crafting a plausible prior distribution, we now assume to have access to i.i.d. samples $\{x_i\}^n_{i=1}$ from the \emph{true} marginal distribution of $\stx$, say for instance, by taking MR scans of randomly sampled people on the street.
We henceforth denote by $p^\star(x)$ the density associated with this marginal distribution, and by $p^\star(\x|\y)$ the corresponding posterior density resulting from Bayes' theorem. We view $p^\star(x)$ and $p^\star(x|y)$ as the \emph{true} prior and posterior, respectively, inasmuch they represent how nature assigns probability mass for $\stx$.
Unfortunately, it is not possible to perform inference directly with $p^\star(\x|\y)$ because the prior is only known through representative samples. Moreover, the lack of an analytical expression for $p^\star(\x)$ means that we cannot guarantee that $p^\star(\x|\y)$ is well-posed and that the quantities of interest exist and are stable w.r.t. perturbations of $\y$, or that the regularity conditions required for efficient gradient-based MCMC computation are satisfied. Learned Bayesian imaging methods use the available samples to construct an approximation of $p^\star(\x|\y)$. 
In the following we focus on theoretically rigorous strategies for this task that are by construction well-posed and amenable to provably convergent computation\footnote{Here we focus on non-asymptotic convergence results that can be applied to a broad class of models. There are other results, e.g., \cite{fletcher2019plug,Pandit2020}, that provide some guarantees for approximate message passing algorithms, but they require stronger assumptions on the forward model and they only hold asymptotically, in a limit where the dimension of $\X$ and $\Y$ diverge in a specific way.}. Specifically, we discuss two strategies: a) \ac{PnP} Bayesian methods that encode the prior distribution in the form of an end-to-end denoising neural network that is used within a Bayesian computation algorithm, and b) algorithms that rely on a generative model trained to reproduce the prior distribution from its samples, such as generative adversarial networks (GANs) or variational auto-encoders (VAEs).

\paragraph{Bayesian methods with plug-and-play priors}
The aim is now to learn a prior by exploiting its relation to a learned denoiser, similarly as in Sec. \ref{sec:pnp}. For that purpose, we denote by $D_\sigma^\star : \X \mapsto \X$ the optimal minimum mean-squared-error (MMSE) denoiser that estimates $\stx$ from its noisy observation $\stu = \stx + \sigma \stz$, where $\stz$ is a standard Gaussian random variable. From Bayesian decision theory, $D_\sigma^\star$ is given by $D_\sigma^\star(u)= \displaystyle\textrm{E}[\stx|\stu=u]$ for any $u \in \X$, where the expectation is computed under the assumption that $\stx$ has marginal density $p^\star$. In practice, $D_\sigma^\star$ is of course unknown, because $p^\star$ is unknown. However, it can be approximated using a deep neural network $D_\sigma$ trained on the available samples from $p^\star$.

Plug-and-play (PnP) Bayesian methods now use this denoiser $D_\sigma$ trained to mimic $D_\sigma^\star$ in order to perform approximate inference w.r.t. the oracle $p^\star(\x|\y)$. More precisely, this can be achieved 
by mimicking gradient-based Bayesian computation algorithms that target a regularized approximation of $p^\star(\x|\y)$, which, by construction, verifies the regularity properties required for fast convergence \cite{laumont2021b,laumont2021a}. In particular, \ac{PnP} Bayesian methods stem from the observation that $D_\sigma^\star$ is related to $p^\star$ by Tweedie's identity: \begin{equation*}
    \sigma^2 \nabla \log p_\sigma^\star(x) = D_\sigma^\star(x)-x,
\end{equation*}
for all $x \in \X$ and $\sigma > 0$,
where $p_\sigma^\star$ is a regularized approximation of $p^\star$ obtained via the convolution of $p^\star$ with a Gaussian smoothing kernel of bandwidth $\sigma$. Unlike $p^\star$ which may be degenerate or non-smooth, $p_\sigma^\star$ is by construction proper and smooth, with its gradient $x \mapsto \nabla \log p_\sigma^\star(x)$ being globally Lipschitz continuous under mild conditions \cite{laumont2021b}. Also, $p_\sigma^\star(x)$ can be made arbitrarily close to $p^\star(x)$ by reducing $\sigma$ to control the approximation error involved.


Equipped with this regularized prior, \ac{PnP} Bayesian methods use Bayes' theorem to derive the regularized posterior density $p_\sigma^\star(\x|\y) \propto p(y|x)p_\sigma^\star(x)$. Under gentle assumptions on the likelihood $p(y|x)$, $p_\sigma^\star(\x|\y)$ inherits the favorable regularity properties of $p_\sigma^\star(x)$ and provides an approximation to $p^\star(\x|\y)$ that is amenable to efficient computation by gradient-based algorithms such as the \ac{ULA} and \ac{SGD} \cite{laumont2021b}. By controlling $\sigma$, the approximation $p_\sigma^\star(\x|\y)$ can be made as close to $p^\star(\x|\y)$ as required in order to control the estimation bias. This, however, comes at the expense of additional computation due to slower convergence of gradient-based algorithms. In addition, \cite{laumont2021b} provides verifiable conditions that guarantee that $p_\sigma^\star(\x|\y)$ is well-posed w.r.t. the TV distance, and key quantities such as the posterior moments exist.

With regards to the maximum a-posteriori probability (MAP) estimation for $p_\sigma^\star(\x|\y)$, defined as
\begin{equation*}
    \hat{x}_{\text{map}} \in \argmax_{x \in \X} \, p_\sigma^\star(\x|\y),
\end{equation*}
it is worth mentioning two main results from \cite{laumont2021a} before discussing practical computational issues. First, MAP solutions of $p_\sigma^\star(\x|\y)$ lie in a neighborhood of MAP solutions of $p^\star(\x|\y)$ and they vary in a stable manner w.r.t. $\sigma$, with the two sets of solutions coinciding as $\sigma \rightarrow 0$. Second, MAP solutions of $p_\sigma^\star(\x|\y)$ are locally Lipschitz continuous w.r.t. to perturbations in $y \in \Y$, which is a weak form of well-posedness (see \cite{laumont2021a} for details).

From a Bayesian computation viewpoint, the main theoretical insight underpinning \ac{PnP} Bayesian methods such as the \ac{PnP}-ULA and \ac{PnP}-SGD studied in \cite{laumont2021b,laumont2021a} is that one can use $\nabla \log p_\sigma^\star$ to formulate idealized \ac{ULA} and \ac{SGD} algorithms for inference w.r.t. $p_\sigma^\star$, and subsequently substitute $\displaystyle\nabla \log p_\sigma^\star(x) = ({D^\star_\sigma(x)-x})/{\sigma^2}$ within these algorithms by the learned approximation $({D_\sigma(x)-x})/{\sigma^2}$ without significantly affecting their convergence properties, even if $D_\sigma(x)$ is not a gradient or a maximally monotone operator. Approximating the oracle denoiser $D_\sigma^\star$ by a learned denoiser $D_\sigma$ introduces some bias; i.e., the algorithms produce a solution in the neighborhood of the oracle solution that would be produced by the idealized algorithms. The magnitude of this bias w.r.t. the oracle depends primarily on how close $D_\sigma$ is to $D_\sigma^\star$.


\begin{mdframed}[style=mystyle,frametitle= The plug-and-play unadjusted Langevin algorithm (PnP-ULA) and stochastic gradient-descent (PnP-SGD)]
PnP-ULA to sample from the regularized posterior $p_\sigma^\star(x|y)$ is defined by the following recursion \cite{laumont2021b}, where $k \in \Natural$ :
\begin{equation}
    \tilde{\stx}_{k+1} = \tilde{\stx}_k + \delta \nabla \log p(y|\tilde{\stx}_k) + \frac{\delta}{\sigma^2} \left[D_\sigma(\tilde{\stx}_k)-\tilde{\stx}_k\right]+ \frac{\delta}{\lambda} \left[\Pi_\mathrm{C}(\tilde{\stx}_k)-\tilde{\stx}_k\right] + \sqrt{2\delta}\,\stz_{k+1}.
\label{eq:pnp_ula_def}
\end{equation}
Here, $\delta$ is a step-size, $\lambda$ is a tail regularization parameter, $\mathrm{C} \subset \X$ denotes a compact convex set that contains most of the prior probability mass of $\stx$, $\Pi_\mathrm{C}$ is the projection operator onto $\mathrm{C}$, and $\{\stz_k \}_{k\in \Natural}$ are i.i.d.\@ standard Gaussian random variables.

For maximum a-posteriori probability (MAP) estimation for $p^\star(x|y)$, the PnP-SGD algorithm is defined by the following recursion \cite{laumont2021a}, for $k \in \Natural$:
\begin{align}
    \tilde{\stx}_{k+1} = &\tilde{\stx}_k + \delta_k \nabla \log p(y|\tilde{\stx}_k) + \frac{\delta_k}{\sigma^2} \left[D_\sigma(\tilde{\stx}_k)-\tilde{\stx}_k\right] + \delta_k \stz_{k+1}\, ,
    \label{eq:pnp_sgd_def}
\end{align}
where $\{\delta_k\}_{k\in \Natural}$ is a family of decreasing positive step-sizes and $\{\stz_k \}_{k \in \Natural}$ are again i.i.d. standard Gaussian random variables.\\ 
Convergence guarantees for these two algorithms require that the denoiser satisfies the same condition as in \eqref{cond_denoiser_pnpDRS}, for all $u,v \in \X$, and for some $\epsilon>0$. In addition, to characterize the error introduced by approximating $D^\star_\sigma$ through $D_\sigma$, one needs that for any $R>0$ there exists $M_R \geq 0$ such that 
$\left\|D_{\sigma}(u)-D^\star_{\sigma}(u)\right\|_2 \leq M_R$ for all $u \in \X$ with $\|u\|_2 < R$. Further, the likelihood $p(y|x)$ is assumed to be finite and differentiable w.r.t. $\x$, with $\nabla \log p(y|x)$ being $L_y$-Lipschitz.\\
Then, for any $\displaystyle\delta < \frac{1}{3}\left(\frac{\epsilon}{\sigma}+L_y+\frac{1}{\lambda}\right)$ and $\lambda \in \left(0,\frac{1}{2}\left(\frac{\epsilon}{\sigma} + 2 L_y\right)\right)$, the Markov chain generated by \eqref{eq:pnp_ula_def} converges geometrically fast to a neighborhood of $p^\star_\sigma(x|y)$, in TV and Wasserstein-1 distances. The asymptotic bias and the convergence rate depend on $\delta,\lambda,\mathrm{C},M_R$ and the dimension of $\X$ (see \cite[Section 3.2]{laumont2021b} for details).\\
For the PnP-SGD algorithm in \eqref{eq:pnp_sgd_def}, assume that the step-sizes satisfy
$$
\lim_{k\rightarrow\infty} \delta_k = 0\, , \quad \sum_{k=0}^\infty \delta_k = + \infty\, , \quad \sum_{k=0}^\infty \delta^2_k < + \infty\, .
$$
Then, it is shown in \cite{laumont2021a} that all \emph{stable} sequences generated by \eqref{eq:pnp_sgd_def} converge to a neighborhood of the stationary points of $\nabla \log p^\star_\sigma(\x|\y)$, and the size of the neighborhood is controlled by $M_R$ (see \cite[Proposition 3]{laumont2021a} for technical details). Whether all the sequences generated by \eqref{eq:pnp_sgd_def} are stable is an open question.
\end{mdframed}

\begin{figure}[ht!]
\centering
  \subfloat[22.62 dB, 0.66]{\includegraphics[width=0.28\linewidth]
  {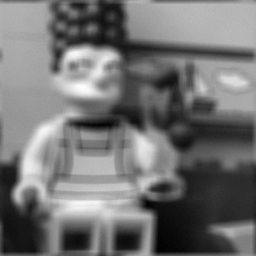}}
  \hspace{0.02in}
  \subfloat[30.62 dB, 0.93]{\includegraphics[width=0.28\linewidth]
  {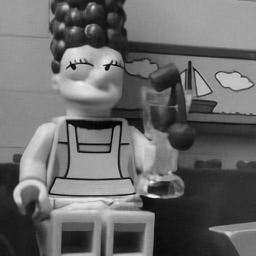}}
  \hspace{0.02in}
  \subfloat[28.90 dB, 0.90]{\includegraphics[width=0.28\linewidth]
  {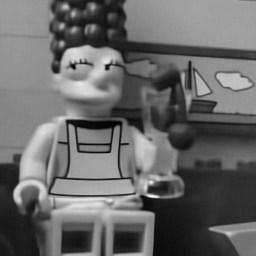}}
  \\
  \subfloat[scale $2\times 2$]{\includegraphics[width=0.28\linewidth]
  {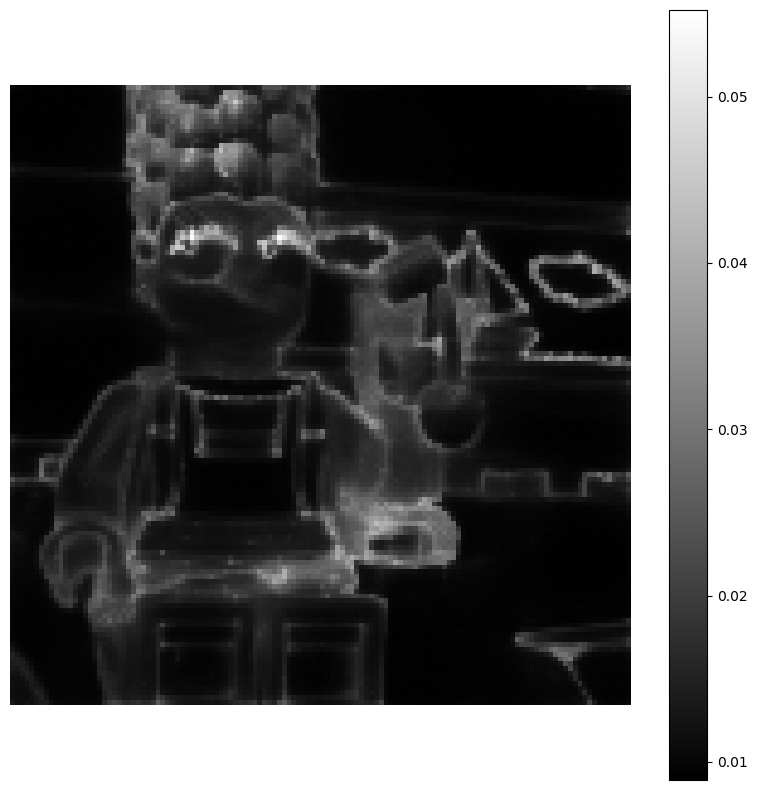}}
  \hspace{0.02in}
  \subfloat[scale $4\times 4$]{\includegraphics[width=0.28\linewidth]
  {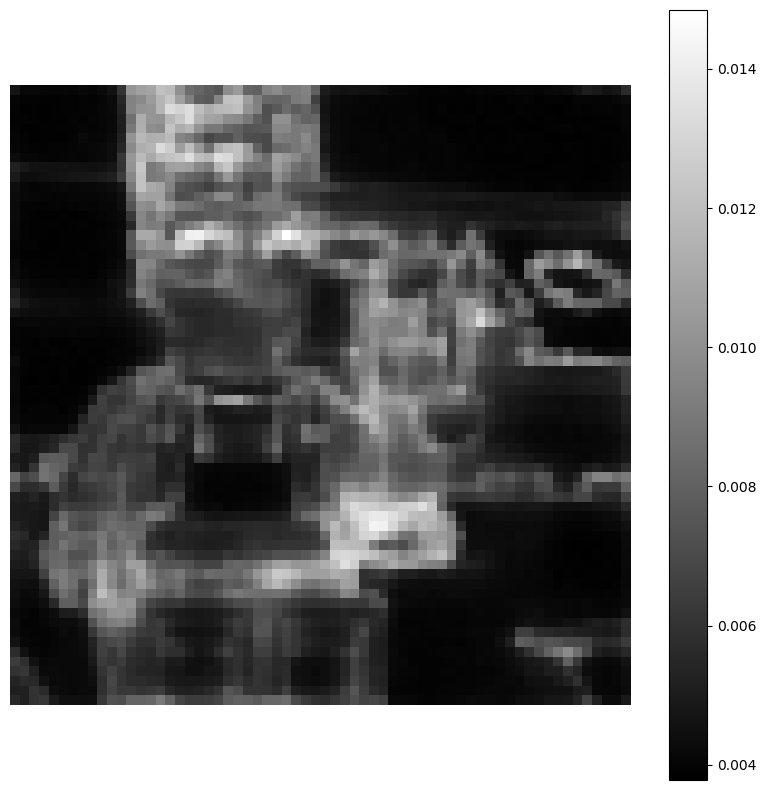}}
  \hspace{0.02in}
  \subfloat[scale $8\times 8$]{\includegraphics[width=0.28\linewidth]
  {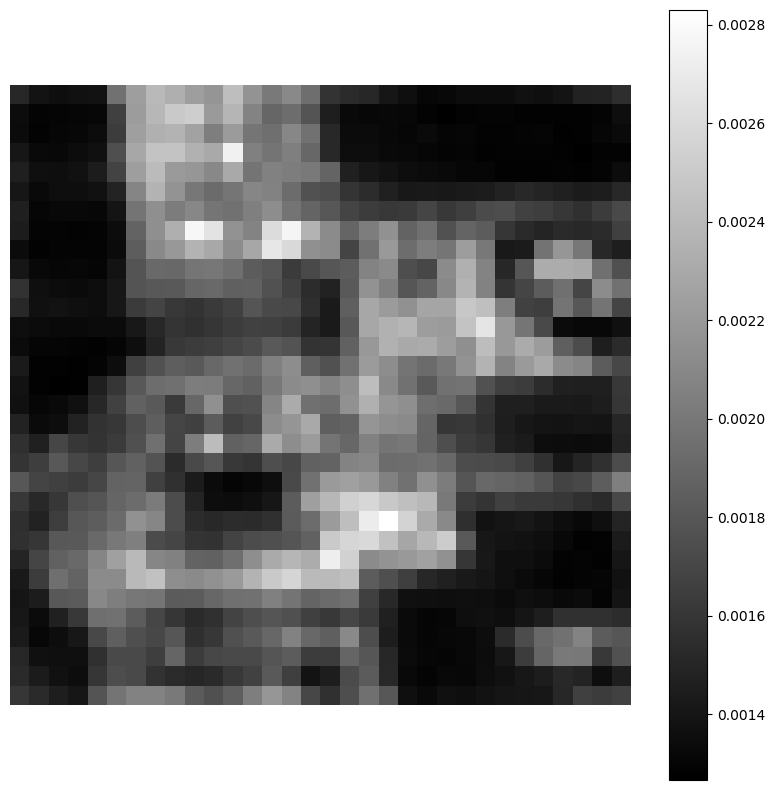}}
\caption{\small{Image deblurring using the denoiser \cite{pmlr-v97-ryu19a} in a \ac{PnP} fashion. Top row shows data (left), a blurred $256\times 256$ image with additive Gaussian noise with standard deviation $1/255$, along with \ac{MMSE} solution from \ac{PnP}-ULA (middle) and \ac{MAP} solution from \ac{PnP}-SGD (right). The corresponding PSNR (dB) and SSIM values are indicated below the images.} The bottom row shows uncertainty plots at the scales of $2\times 2$ (left), $4\times 4$ (middle), and $8\times 8$ pixels (right), computed by \ac{PnP}-ULA. See \cite{laumont2021b} for details.}
\label{fig:PnPBayes_results}
\end{figure}

For illustration, Fig.~\ref{fig:PnPBayes_results} presents the results of an image deblurring experiment where \ac{PnP}-ULA and \ac{PnP}-SGD are used to compute the \ac{MMSE} and \ac{MAP} 
estimators, respectively,
using the denoiser of \cite{pmlr-v97-ryu19a} with $\sigma = 5/255$. The main strength of Monte Carlo algorithms such as \ac{PnP}-ULA is their capacity for Bayesian analyses beyond point estimation, such as uncertainty quantification, as depicted in the bottom row of Fig.~\ref{fig:PnPBayes_results}.

\paragraph{Bayesian methods with generative priors} A prominent alternative to plug-and-play denoising is to construct a prior distribution for $\stx$ by leveraging recent developments in deep generative modeling, such as \acp{VAE}, \acp{GAN}, and normalizing flows. The Bayesian approach in \cite{holden2021bayesian} adopts the manifold hypothesis to construct a prior distribution that is supported on a sub-manifold of $\X$, the dimension of which is much smaller than the ambient dimension of $\X$. Operating on this manifold of dramatically reduced dimensionality simultaneously allows to effectively regularize the posterior and perform computations efficiently. The reduction in dimensionality is achieved by introducing a latent Gaussian variable $\stz \sim \mathcal{N}(0,\mathbb{I}_d)$ on $\Real^d$ and a deep generative network $\Op{G}_\theta \colon \Real^d \mapsto \X$ designed such that the random variable $\Op{G}_\theta(\stz)$ is close in distribution to $\stx$. In \cite{holden2021bayesian}, $\Op{G}_\theta$ is derived from a VAE architecture, but the strategy is agnostic to the choice of the specific generative model, and a GAN or a normalizing flow-based model could be used instead. Given this construction, the latent posterior is derived as $p(\z|\y)\propto p(\y|\z)p(\z)$, where $p(\z)$ is a standard Gaussian density on $\Real^d$ and the likelihood is $p(\y|\z) = p_{\sty|\stx}(\y|\Op{G}_\theta(\z))$. The posterior distribution of $(\stx|\sty=\y)$ is then given by using $\Op{G}_\theta$ to map $(\stz|\sty=\y)$ onto $\X$. The resulting Bayesian inversion is shown to be well-posed under mild conditions on the likelihood, and important quantities such as the posterior mean exist.

\begin{mdframed}[style=mystyle,frametitle=Well-posedness of linear Bayesian inverse problems with generative priors]
Consider the generative model $\stx = \Op{G}_\theta(\stz)$ where $\stz \sim \mathcal{N}(0,\mathbb{I}_d)$ is a latent random variable and $\Op{G}_\theta \colon \Real^d \mapsto \X \subseteq \mathbb{R}^p$ is a Lipschitz-continuous neural network. Suppose that the observed data $y\in \mathbb{R}^n$ is a realization of a random variable $\sty = \ForwardOp\Op{G}_\theta(\stz) + \ste$, where $\ForwardOp \in \mathbb{R}^{n\times p}$ and $\ste$ is Gaussian noise. Then, it is shown in \cite{holden2021bayesian} that the posterior distributions associated with $(\stz|\sty=\y)$ and $(\stx|\sty=\y)$ are well-posed w.r.t. the TV and Wasserstein-2 distances. Moreover, all posterior moments exist, and in particular the minimum mean-squared error (MMSE) Bayesian estimators for $(\stz|\sty=\y)$ and $(\stx|\sty=\y)$ are well-posed.
\end{mdframed}


With regards to computation, \cite{holden2021bayesian} takes advantage of the fact that $\stz$ is relatively low-dimensional and has a Gaussian prior to generate samples for $(\stz|\sty=\y)$ by using the following simple MCMC procedure specialized for this class of models: for any $k \in \Natural$, draw $\stz^\dagger \sim \mathcal{N}(\sqrt{1-\delta^2}\stz_k,\mathbb{I}_d)$ and set $\stz_{k+1} = \stz^\dagger$ with probability $\min(1,p(\y|\stz^\dagger)/p(\y|\stz_k))$; otherwise set $\stz_{k+1} = \stz_{k}$. This MCMC algorithm, known as the preconditioned Crank-Nicolson algorithm, is  provably convergent under mild assumptions on $p(\y|\z)$ that are verified in particular for linear Gaussian observation models of the form $\sty = \ForwardOp\Op{G}_\theta(\stz) + \ste$. Given Monte Carlo samples for the latent variable $(\stz|\sty=\y)$, samples for $(\stx|\sty=y)$ are obtained by applying the generator $\Op{G}_{\theta}$.



A Bayesian model with a prior encoded by a \ac{VAE} was considered in \cite{gonzalez2021solving}, with an objective of \ac{MAP} estimation. Unlike the approach in \cite{holden2021bayesian}, where one constraints $\stx$ to take values in the range of $\Op{G}_\theta$ in order to reduce dimensionality, \cite{gonzalez2021solving} considers an augmented model $p(\x,\z)$ on $\X \times \Real^d$ that concentrates mass in the neighborhood of $\x = \Op{G}_\theta(\z)$ while carefully allowing for deviations from this sub-manifold to better fit the training images. This leads to an augmented posterior distribution $p(\x,\z|\y)\propto p(\y|\x)p(\x,\z)$ that is a more accurate model than the marginal posterior model considered in \cite{holden2021bayesian}. Inference with $p(\x,\z|\y)$ is significantly more computationally challenging, a difficulty that was addressed by focusing exclusively on \ac{MAP} estimation. Crucially, the authors established that the potential $(\x,\z) \mapsto -\log p(\x,\z | \y)$ is weakly bi-convex under realistic conditions on the \ac{VAE}, and subsequently proposed three provably convergent alternating optimization schemes to compute a critical point of this potential efficiently. 

\section{Conclusions and outlook}
\label{sec:conclusion}
In scientific disciplines where imaging drives new discoveries or in real-world applications where imaging is used for making critical decisions, it is essential to have mathematical correctness guarantees for the algorithms used for image recovery. While the classical variational approaches come with such certificates, they fall short in terms of empirical performance as compared to the modern data-driven imaging algorithms. We formalized different notions of correctness as they apply to image reconstruction methods and surveyed some of the notable deep learning-based approaches, both deterministic and stochastic, that fit within these notions. We discussed some of the essential components, e.g., network architecture design, training strategies, etc. that typically aid deriving such theoretical certificates. While we sought to dispel the widely held belief about the black-box nature of deep learning algorithms for image reconstruction, we also highlighted the gaps in theoretical understanding about well-performing methods rooted in robust heuristics. The methods we reviewed in this article broadly derive their origin from the variational regularization framework and convex analysis, two of the major theoretical pillars that the classical methods rest on. In fact, convexity arose as a recurring theme for proving convergence results in both deterministic and stochastic settings, which underscores the importance of \acp{ICNN} for combining classical theory with data-driven learning. In summary, we argued that the classical mathematical machinery can go a long way when it comes to devising and analyzing data-driven methods, leading to better reliability and transparency of deep learning for imaging.         

\bibliographystyle{IEEEtran}
\bibliography{ref}

 \begin{IEEEbiography}[{\includegraphics[height=1.0in,clip,keepaspectratio]{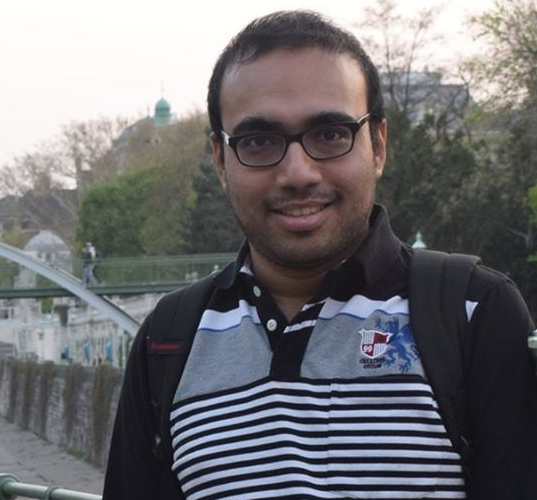}}]{Subhadip Mukherjee} is currently a Lecturer (Assistant Professor) of machine learning and AI at the Department of Computer Science, University of Bath, UK. Prior to this, he held two postdoctoral positions, at the Department of Applied Mathematics and Theoretical Physics, University of Cambridge, UK (2020-2022); and at the Department of Mathematics, KTH, Sweden (2018-2020). He completed his Ph.D. in 2018 from the Department of Electrical Engineering, Indian Institute of Science, Bangalore, specializing in sparsity-regularized inverse problems. His research interests include areas at the interface of machine learning, inverse problems, convex optimization, and statistics. Specifically, he is interested in developing novel machine learning algorithms for computational imaging problems.
 \end{IEEEbiography}

 \begin{IEEEbiography}[{\includegraphics[height=1.0in,clip,keepaspectratio]{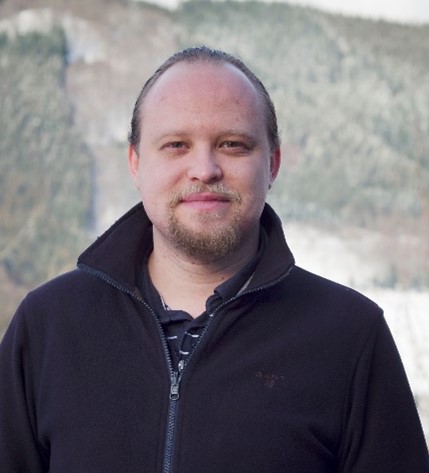}}]{Andreas Hauptmann} is Academy Research Fellow and Associate Professor of Computational Mathematics at the Research Unit of Mathematical Sciences, University of Oulu, Finland. He has worked prior to this as Research Associate at the Department of Computer Science, University College London, UK. He received his Ph.D. in 2017 in Applied Mathematics from the University of Helsinki, Finland. His research interest is in combining model-based inversion techniques with data driven methods for tomographic reconstructions.
 \end{IEEEbiography}

 \begin{IEEEbiography}[{\includegraphics[height=1.0in,clip,keepaspectratio]{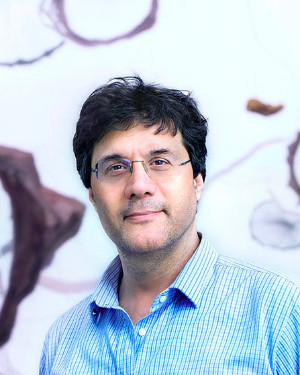}}]{Ozan \"Oktem} is a Professor in Computational Science at the Department of Information Technology, Uppsala University and Associate Professor in Numerical Analysis at the Department of Mathematics, KTH - Royal Institute of Technology,  Stockholm, Sweden. He received his Ph.D. in 1999 in Mathematics from Stockholm University, Sweden. He worked for 13 years in industry before returning to academia in 2009. His recent focus is on combining model based approaches with \acp{DNN} for uncertainty quantification and task adapted reconstruction in large scale inverse problems, with concrete challenges in imaging applications from various scientific fields.
\end{IEEEbiography}
\begin{IEEEbiography}[{\includegraphics[height=1.0in,clip,keepaspectratio]{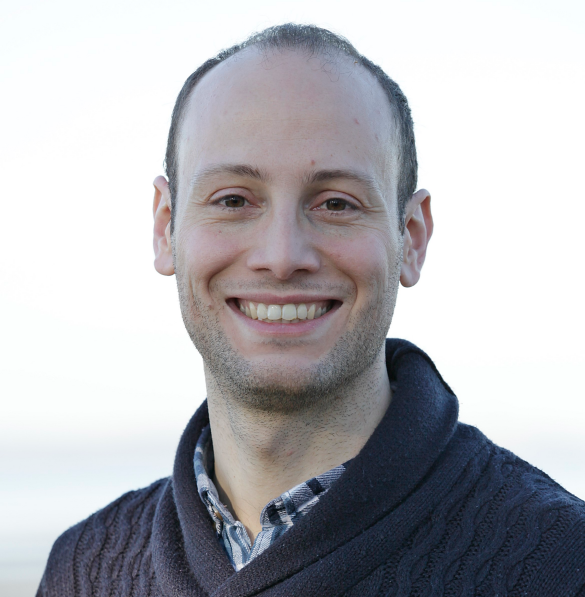}}]{Marcelo Pereyra} is an Associate Professor at the Maxwell Institute for Mathematical Sciences and the School of Mathematical and Computer Sciences, Heriot-Watt University. He 
obtained a Ph.D. degree in Signal Processing from the University of Toulouse in 2012 and was a Research Fellow in Statistics at the University of Bristol (2012 - 2016), funded by a Marie Curie Intra-European Fellowship for Career Development, a Brunel Postdoctoral Research Fellowship in Statistics, and a Postdoctoral Research Fellowship from French Ministry of Defence.
In 2019 he was an Invited Professor at the Institut Henri Poincaré in Paris during the "Mathematics of Imaging" trimester. His research focuses on new Bayesian statistical theory, methodology and algorithms to solve challenging inverse problems related to computational imaging. 
 \end{IEEEbiography}
 
 \begin{IEEEbiography}[{\includegraphics[height=1.0in,clip,keepaspectratio]{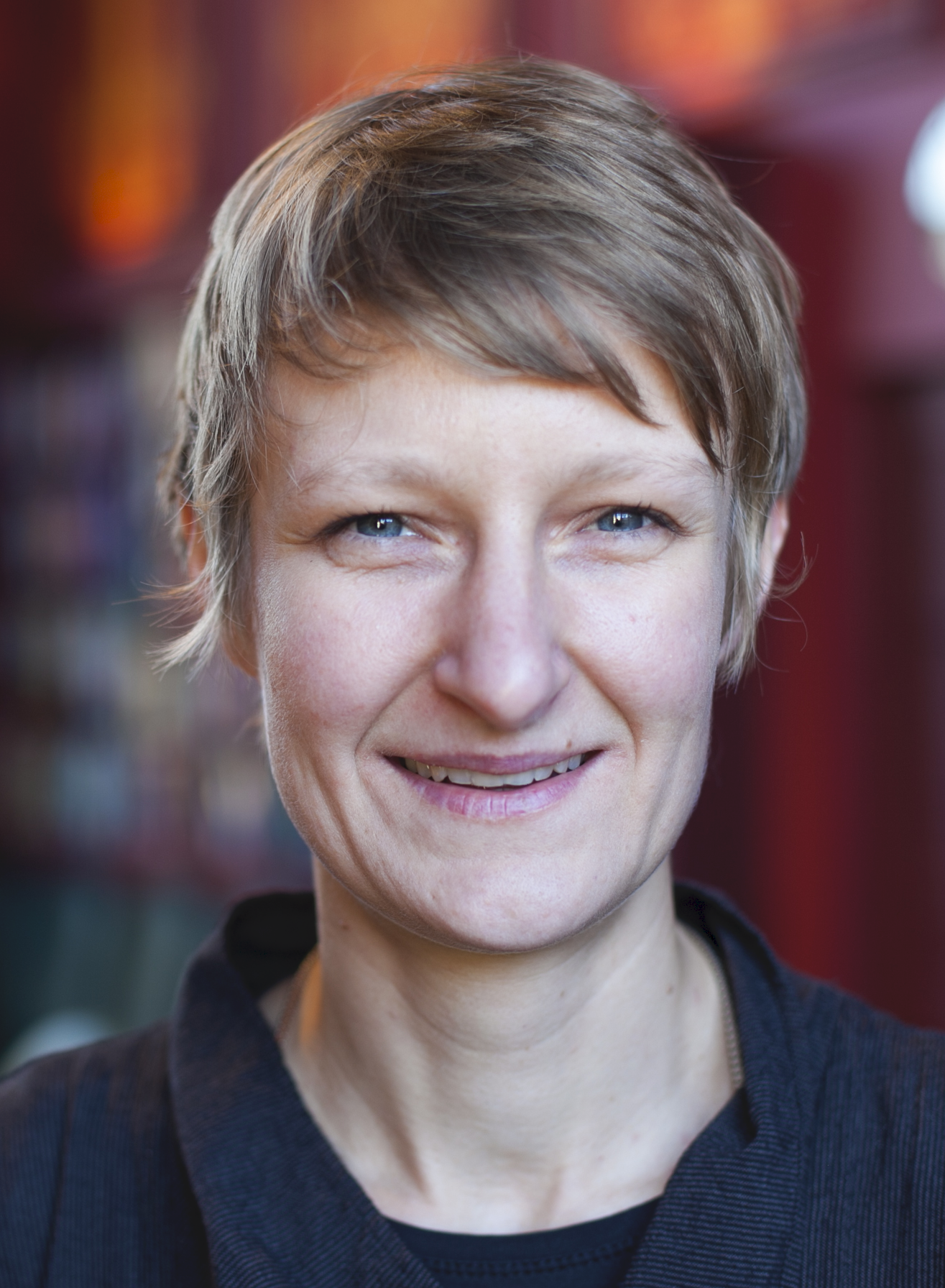}}]{Carola-Bibiane Sch\"onlieb} is Professor of Applied Mathematics at the University of Cambridge, where she is head of the Cambridge Image Analysis group and co-Director of the EPSRC Cambridge Mathematics of Information in Healthcare Hub. Carola graduated from the Institute for Mathematics, University of Salzburg (Austria) in 2004. From 2004 to 2005 she held a teaching position in Salzburg. She received her PhD degree from the University of Cambridge (UK) in 2009. After one year of postdoctoral activity at the University of Göttingen (Germany), she became a Lecturer at Cambridge in 2010, promoted to Reader in 2015 and promoted to Professor in 2018. Since 2011 she is a fellow of Jesus College Cambridge and since 2016 a fellow of the Alan Turing Institute, London. Her current research interests focus on variational methods, partial differential equations and machine learning for image analysis, image processing and inverse imaging problems.
 \end{IEEEbiography}
\noindent


\end{document}